\documentclass[lettersize,journal]{IEEEtran}
\usepackage{amsmath,amsfonts}
\usepackage{algorithmic}
\usepackage{algorithm}
\usepackage{array}
\usepackage[caption=false,font=footnotesize,labelfont=rm,textfont=rm]{subfig}
\usepackage{textcomp}
\usepackage{stfloats}
\usepackage{url}
\usepackage{verbatim}
\usepackage{graphicx}
\usepackage{multirow}
\usepackage{cite}
\usepackage{hyperref}
\usepackage{color}

\begin{document}

\title{MSSF: A 4D Radar and Camera Fusion Framework With Multi-Stage Sampling for 3D Object Detection in Autonomous Driving}

\author{Hongsi Liu, $\text{Jun Liu}^\dag$,~\IEEEmembership{Senior Member,~IEEE}, Guangfeng Jiang, Xin Jin,~\IEEEmembership{Member,~IEEE}
\thanks{Hongsi Liu, Jun Liu, and Guangfeng Jiang are with the Department of Electronic Engineering and Information Science, University of Science and Technology of China, Hefei 230027, China (e-mail: liuhs3@mail.ustc.edu.cn; junliu@ustc.edu.cn; jgf1998@mail.ustc.edu.cn).}
\thanks{Xin Jin is with Ningbo Institute of Digital Twin, Eastern Institute of Technology, Ningbo, Zhejiang 315201, China (e-mail: jinxin@eitech.edu.cn).}
\thanks{$^\dag$Corresponding author.}
}

\markboth{Journal of \LaTeX\ Class Files}%
{Liu \MakeLowercase{\textit{et al.}}: MSSF: A 4D Radar and Camera Fusion Framework With Multistage Sampling for 3D Object Detection}


\maketitle

\begin{abstract}
As one of the automotive sensors that have emerged in recent years, 4D millimeter-wave radar has a higher resolution than conventional 3D radar and provides precise elevation measurements. But its point clouds are still sparse and noisy, making it challenging to meet the requirements of autonomous driving. Camera, as another commonly used sensor, can capture rich semantic information. As a result, the fusion of 4D radar and camera can provide an affordable and robust perception solution for autonomous driving systems. 
However, previous radar-camera fusion methods have not yet been thoroughly investigated, resulting in a large performance gap compared to LiDAR-based methods. Specifically, they ignore the feature-blurring problem and do not deeply interact with image semantic information. 
To this end, we present a simple but effective multi-stage sampling fusion (MSSF) network based on 4D radar and camera. On the one hand, we design a fusion block that can deeply interact point cloud features with image features, and can be applied to commonly used single-modal backbones in a plug-and-play manner. The fusion block encompasses two types, namely, simple feature fusion (SFF) and multi-scale deformable feature fusion (MSDFF). The SFF is easy to implement, while the MSDFF has stronger fusion abilities.
On the other hand, we propose a semantic-guided head to perform foreground-background segmentation on voxels with voxel feature re-weighting, further alleviating the problem of feature blurring. Extensive experiments on the View-of-Delft (VoD) and TJ4DRadset datasets demonstrate the effectiveness of our MSSF. Notably, compared to state-of-the-art methods, MSSF achieves a 7.0\% and 4.0\% improvement in 3D mean average precision on the VoD and TJ4DRadSet datasets, respectively.
It even surpasses classical LiDAR-based methods on the VoD dataset.
Code is available at \href{https://github.com/EricLiuhhh/MSSF.git}{https://github.com/EricLiuhhh/MSSF.git}.
\end{abstract}


\section{Introduction}\label{sec:intro}
\IEEEPARstart{A}UTONOMOUS driving is a hot topic in both academia and industry in recent years, and the research around it can be mainly divided into three parts, namely perception, planning \& decision, and control\cite{mao3DObjectDetection2023a}. Perception plays an important role in autonomous driving. It covers a lot of content, such as object detection\cite{langPointPillarsFastEncoders2019a,zhouVoxelNetEndtoEndLearning2018,shiPointRCNN3DObject2019a}, tracking\cite{weng2020AB3DMOT}\cite{pang2022simpletrack}, and segmentation\cite{jiang2024mwsis}\cite{zhang2024spatial}. 3D object detection, as one of the main tasks in perception, has attracted much attention nowadays. Its purpose is to obtain the categories and 3D bounding boxes of critical objects (e.g., cars and pedestrians) in 3D scenes from sensor data. To achieve this, a wide variety of sensors can be used, such as LiDAR, radar, and camera. 


LiDAR can obtain high-precision point clouds of 3D scenes which well reflect the geometric information of objects, thereby achieving remarkable detection performance.
However, it may be infeasible due to the high cost and susceptibility to adverse weather conditions such as rain and fog.

Millimeter wave radar (refer to radar for convenience) is a more common vehicle sensor\cite{karangwa2023vehicle} compared to LiDAR, with the advantages of low price, slight influence by rain and fog, and long detection range\cite{zhouDeepRadarPerception2022}. Before the emergence of 4D radar, conventional 3D radar, which can measure distance, azimuth, and Doppler, is primarily employed. Nevertheless, the lack of elevation measurements limits the perception capability of 3D radar. 4D radar addresses this limitation and offers a high resolution\cite{zhouDeepRadarPerception2022}. As a result, it can provide relatively dense three-dimensional point clouds like LiDAR, which is increasingly being recognized as an affordable alternative to LiDAR.


Although 4D radar point clouds share numerous similarities with LiDAR point clouds, they are still sparse and noisy due to the limited ranging and angle resolution, multipath effects, and penetrability\cite{zhouDeepRadarPerception2022}. 
Hence, relying solely on 4D radar point clouds for 3D object detection has limitations in performance. However, there is substantial potential for performance improvement through fusion with other sensors.  
Meanwhile, it is worth noting that cameras are relatively cheap and easy to deploy, which can provide rich semantic information due to the high spatial resolution and ability to perceive the color and texture of objects.

\begin{figure}[!t]
    \centering
    \includegraphics[width=3.45in]{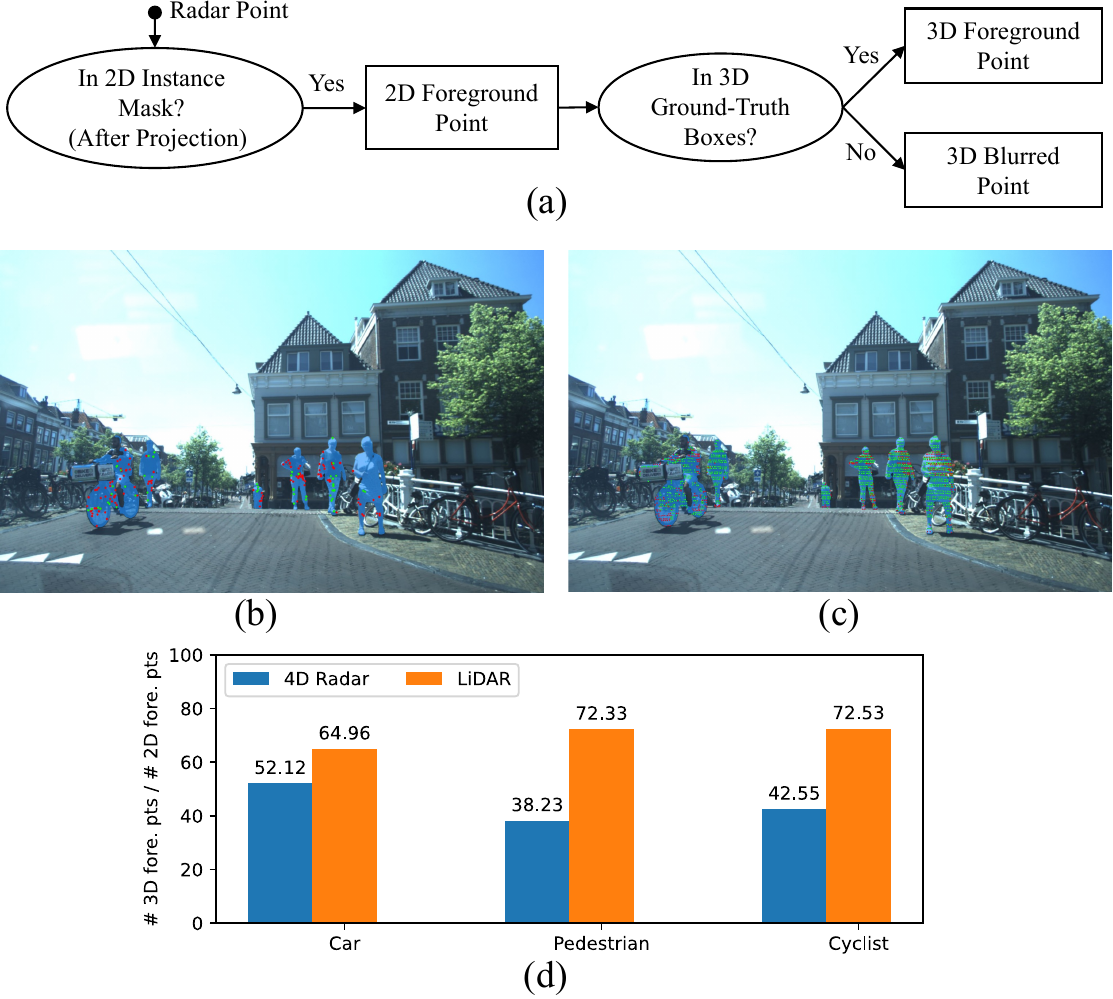}
    \caption{An explanation of the feature-blurring problem. (a) elucidates the definitions of 2D foreground points, 3D foreground points, and 3D blurred points.
    (b) and (c) show the radar points and LiDAR points projected onto the image, respectively. The blue mask is the instance segmentation generated by SAM\cite{kirillovSegmentAnything2023}, the green points represent the 3D foreground points, and the red points represent the 3D blurred points. 
    (d) illustrates quantitatively by averaging the ratio of the number of 3D foreground points to the number of 2D foreground points over around 200 instance masks for each class. ``\# 3D fore. pts'' represents the number of 3D foreground points, and ``\# 2D fore. pts'' represents the number of 2D foreground points.}
    \label{fig:motivate}
\end{figure}

Considering the above factors, some recent works attempt to integrate information from both radar and camera modalities. Some researchers\cite{kimCRNCameraRadar2023a,nabatiCenterFusionCenterbasedRadar2021,kimCRAFTCameraRadar3D2023} specifically design the fusion strategy of 3D radar point clouds and images, and have shown obvious improvement over single-modal baselines. 
As an increasing number of 4D radar datasets\cite{meyer2019automotive,palffyMultiClassRoadUser2022,zhengTJ4DRadSet4DRadar2022,paek2022k} are released, recent studies\cite{yaoRadarCameraFusionObject2023}\cite{xiongLXLLiDARExcluded2023b} focus on the fusion method for 4D radar and camera. RCFusion\cite{yaoRadarCameraFusionObject2023} adopts orthographic feature transform (OFT)\cite{roddick2018orthographic} to obtain image bird's eye view (BEV) feature maps and design an interactive attention module to fuse the BEV feature maps of point clouds and images. LXL\cite{xiongLXLLiDARExcluded2023b} uses a depth-based sampling strategy to lift image features to 3D space with the help of the predicted radar occupancy. 

Although existing 4D radar and camera fusion methods achieve good performance, they are mostly based on the BEV domain fusion framework designed for LiDAR-camera fusion,  without explicitly considering the characteristics of radar.
Consequently, the potential of radar-camera fusion has not been fully explored, leading to a large performance gap compared to LiDAR-based methods.
Specifically, in contrast to the dense point clouds produced by LiDAR, the 4D radar captures fewer details in terms of geometry information. 
Hence, radar-camera fusion places a greater reliance on image semantic information compared to LiDAR-camera fusion. However, existing methods only fuse radar and camera information on the BEV, lacking sufficient feature interaction and neglecting the characteristic that radar relies more on semantic information from images.

Meanwhile, serious feature-blurring problems may arise when projecting radar points onto the corresponding image for feature sampling. In Fig. \ref{fig:motivate}, we explain this problem through visualization and statistics in the View-of-Delft (VoD) dataset. We first provide the definitions of 2D foreground points, 3D foreground points, and 3D blurred points, as illustrated in Fig. \ref{fig:motivate}(a). A radar point is a 2D foreground point when its corresponding projection point falls within 2D instance masks, because it can sample image features of foreground objects. Further, if a 2D foreground point also falls within 3D ground truth boxes, it is defined as a 3D foreground point, otherwise, it is classified as a 3D blurred point. Fig. \ref{fig:motivate}(b) and (c) show the 2D foreground points of radar and LiDAR, respectively, where the 3D foreground points are marked in green and the 3D blurred points in red. Additionally, as the VoD dataset does not provide segmentation labels, we generate around 200 instance masks of each class with Segment Anything Model (SAM)\cite{kirillovSegmentAnything2023} as exemplified by the blue mask shown in Fig. \ref{fig:motivate}(b) and (c). Interestingly, in 3D object detection, we mainly care about the 3D foreground points, and other points should be regarded as the background points. However, when sampling features on the image, all 2D foreground points capture features of foreground objects, which may lead to false alarms. We refer to this as the feature-blurring problem. 

For quantitative analysis, we further calculate the ratio of the number of 3D foreground points to the number of 2D foreground points for each instance. The result after averaging all instances of each class is shown in Fig. \ref{fig:motivate}(d). It is observed that radar has significantly lower ratios than LiDAR, especially for the pedestrian and cyclist categories. As for the car category, the gap between the two modalities is relatively small, due to the metal materials which weaken the penetration capability of radar. 
This observation indicates that the feature-blurring problem is severe under radar modality.

To address the above problems, we propose a simple but effective multi-stage sampling fusion (MSSF) network based on 4D radar and camera. We fuse point cloud and image features more deeply in the backbone rather than using a separate image BEV branch. In particular, two types of fusion blocks based on image feature sampling are proposed to replace the blocks in some commonly used 3D sparse backbones. Through the deep interaction of point cloud features and image features achieved by the fusion blocks, MSSF can well identify 3D foreground points, thereby alleviating the feature-blurring problem. The cascading of multiple fusion blocks effectively leverages image features, further enhancing detection performance.
Furthermore, we add a semantic-guided head to explicitly help the network distinguish 3D foreground points. Our approach can be easily applied to many single-modal 3D object detection networks based on voxel or pillar.

Our contributions are summarized in four folds:

\begin{enumerate}
    \item Taking into account the sparsity of 4D radar point clouds and the feature-blurring problem, we propose a simple but effective MSSF network. As an early attempt in the field, it provides a strong baseline for later research.
    \item We propose two general, plug-and-play voxel-image feature fusion blocks and insert them into some commonly used 3D sparse networks in multiple stages to achieve deep interaction between voxel and image features. It has good scalability and can be traded off as needed. 
    \item A semantic-guided head is proposed to further alleviate the feature-blurring problem. On the one hand, the segmentation loss guides the network to distinguish between foreground and background points. On the other hand, the segmentation scores are used to re-weight the voxel features to play the role of attention.
    \item Experiments on the VoD and TJ4DRadset datasets show that the proposed method outperforms state-of-the-art radar-camera fusion methods by 7.0\% mAP and 4.0\% mAP, respectively. In particular, for the car category in the VoD dataset, our method achieves a substantial increase of 18.6\% AP compared to state-of-the-art methods. Notably, our MSSF even surpasses some classic LiDAR-based models.
\end{enumerate}

The remainder of this article is organized as follows. Section \ref{sec:related_work} briefly reviews recent works on single-modal and multi-modal 3D object detection. Section \ref{sec:method} introduces our proposed model in detail. 
In Section \ref{sec:exps}, the experimental setups and implementation details are introduced, and the performance of our method is shown and analyzed. Finally, our work is summarized in Section \ref{sec:conclusion}.

\section{Related Work}
\label{sec:related_work}

\subsection{3D Object Detection based on LiDAR/Radar Point Cloud}
LiDAR-based 3D object detection methods can be divided into\cite{mao3DObjectDetection2023a} point-based\cite{shiPointRCNN3DObject2019a}\cite{yang3DSSDPointBased3D2020}, pillar-based\cite{langPointPillarsFastEncoders2019a}, and voxel-based methods\cite{zhouVoxelNetEndtoEndLearning2018}\cite{yanSECONDSparselyEmbedded2018a}\cite{chenVoxelNeXtFullySparse2023a}, etc., according to the representation of point cloud during network processing. In \cite{palmer2023reviewing}, it is found that the point-based methods are less effective for 3D object detection under radar modality compared with pillar-based and voxel-based methods. At present, the 3D object detection methods of radar point clouds are mostly improved based on pillar-based methods. RPFA-Net\cite{xuRPFANet4DRaDAR2021a} replaces the pillar feature extractor in PointPillars\cite{langPointPillarsFastEncoders2019a} with its proposed self-attention-based feature extraction layer, so that context information can be better perceived when encoding pillar features. RCFusion\cite{yaoRadarCameraFusionObject2023} believes that the Doppler and radar cross section (RCS) information of 4D radar is important, and codes the spatial, velocity, and RCS features separately when extracting pillar features. SMURF\cite{liuSMURFSpatialMultiRepresentation2023a} adds additional kernel density estimation features to the backbone. Both MVFAN\cite{yan2023mvfan} and MUFASA\cite{peng2024mufasa} leverage BEV and cylindrical coordinate views simultaneously to better capture radar point cloud features. RadarPillar\cite{musiat2024radarpillars} applies self-attention to pillar features to enlarge the receptive field.

\subsection{3D Object Detection based on LiDAR-Camera Fusion}
LiDAR and camera fusion strategies can be mainly divided into three categories, i.e., early fusion, middle fusion, and late fusion\cite{wang2023multi}. 
In early fusion\cite{voraPointPaintingSequentialFusion2020b, yinMultimodalVirtualPoint2021a, wangPointAugmentingCrossModalAugmentation2021, sindagiMVXNetMultimodalVoxelNet2019a}, image information is embedded into the point cloud in various ways before the point cloud is input to the detection network, which is direct to implementation but lacks deep interaction with point cloud features and image features. For instance, PointAugmenting\cite{wangPointAugmentingCrossModalAugmentation2021} and MVX-Net\cite{sindagiMVXNetMultimodalVoxelNet2019a} both sample image features at a very early stage and adopt simple fusion strategies.
In addition, middle fusion is more effective and has been studied more in recent years\cite{jiaoMSMDFusionFusingLiDAR2023a,liangBEVFusionSimpleRobust,liuBEVFusionMultiTaskMultiSensor2023,liLoGoNetAccurate3D2023a, huang2020epnet,yang2022deepinteraction}. It realizes the interaction and fusion of point cloud features and image features at the feature level. Among these methods, we noted that LoGoNet\cite{liLoGoNetAccurate3D2023a} also projects the centroids of non-empty voxels into the image and uses multi-scale deformable cross-attention to fuse features. However, the features after fusion are only used for the refinement of proposals. The detection results are limited by the quality of proposals generated by the single-modal detector in the first stage. Additionally, EPNet\cite{huang2020epnet} and DeepInteraction\cite{yang2022deepinteraction} also fuse image features through projection. The former emphasizes point-level correspondence and mainly focuses on point-based backbones. The latter operates on point cloud BEV and image planes, and the correspondence between the two modalities is relatively coarse. Compared with these methods, we establish the correspondence between voxels of different resolutions and images from fine to coarse.
As for the late fusion methods\cite{pang2020clocs} which operate on the output of a LiDAR-based 3D object detector and a camera-based 2D object detector. Insufficient feature interactions limit the potential of such methods. The existing research on LiDAR and camera fusion is of significant reference value when studying radar and camera fusion, since both 4D radar and LiDAR data can be presented in the form of point clouds. 

\subsection{3D Object Detection based on Radar-Camera Fusion}
Some studies have pioneered the fusion of 3D radar and camera. CenterFusion\cite{nabatiCenterFusionCenterbasedRadar2021} extends radar points into pillars, which are associated with the proposals predicted from images in the view frustum and assist in 3D attribute estimation. CRAFT\cite{kimCRAFTCameraRadar3D2023} employs the spatio-contextual fusion transformer to refine image proposals by radar measurements. CRN\cite{kimCRNCameraRadar2023a} adopts the BEV fusion framework and utilizes radar occupancy map to assist image view transformation and BEV feature fusion. 
Until recently, some 4D radar datasets\cite{meyer2019automotive, palffyMultiClassRoadUser2022, zhengTJ4DRadSet4DRadar2022, paek2022k} have been released, and there are relatively few studies focusing on the fusion of 4D radar and camera. 
RCFusion\cite{yaoRadarCameraFusionObject2023} employs OFT\cite{roddick2018orthographic} along with its shared attention encoder to generate image BEV feature maps. These feature maps are then fused with radar BEV features in an attention-based manner. LXL\cite{xiongLXLLiDARExcluded2023b} utilizes a sampling-based method to lift image features, in which the radar occupancy grid predicted from the radar BEV feature map and the image depth prediction are used to assist the image BEV features generation. 
UniBEVFusion\cite{zhao2024unibevfusion} proposes Radar Depth Lift-Splat-Shoot, which incorporates additional radar data into the depth prediction process.

These methods are mainly based on the BEV domain fusion framework without explicitly considering the characteristics of radar, resulting in a large performance gap compared to LiDAR-based methods. The sparsity and noisiness of radar point clouds, along with the ill-posed nature of image depth estimation, can cause challenges in these methods when converting image features to voxel or BEV features.
In this study, we propose a new fusion network. Specifically, we directly adopt radar points to sample image features, rather than doing explicit view transformation. By employing an effective multi-stage fusion strategy along with a semantic-guided head, multi-scale image features are fully utilized to achieve comprehensive and efficient fusion with point cloud features.
Even so, our network remains simple and compatible while maintaining high performance.

\section{Proposed Method}
\label{sec:method}

\begin{figure*}[!t]
\centering
\includegraphics[width=6in]{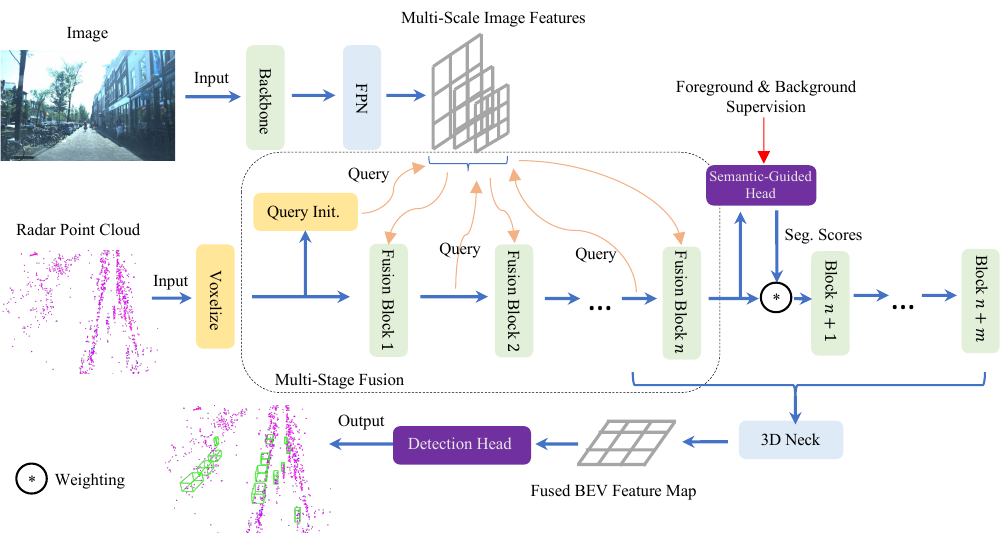}
\caption{The overall architecture of our MSSF. The image branch extracts features from images to obtain multi-scale feature maps. The voxel-image fusion backbone contains $n$ fusion blocks and $m$ ordinary blocks (abbreviated as ``Block'' in the figure), which absorb features from the image feature maps in multiple stages through our proposed fusion blocks. The non-empty voxel features of the last layer fusion block are fed into the semantic-guided head for foreground and background prediction, and the segmentation scores are utilized to weight the voxel features. The multi-scale features output by the last few blocks are passed through the 3D neck to obtain the fused BEV feature map which is sent to the detection head to obtain the final detection results.}
\label{fig:overall}
\end{figure*}

\subsection{Overall Architecture}
The overall network structure is shown in Fig. \ref{fig:overall}, which can be divided into four parts: image branch, voxel-image fusion backbone, semantic-guided head, and detection head. 
\begin{enumerate}
\item The image branch is employed to extract multi-scale features of images and encode image semantic information. In our method, we have no restrictions on the specific structure of the image branch.
\item The voxel-image fusion backbone is one of the key components of the proposed method, which is responsible for extracting point cloud features and deeply fusing with image features output by the image branch. It consists of a cascade of several proposed fusion blocks and ordinary blocks. Through multi-stage fusion, point cloud features are deeply interacted with image features. The 3D neck aggregates the output of the last few blocks, obtaining the fused BEV feature map which contains information from both modalities.
\item The semantic-guided head performs foreground and background segmentation on non-empty voxels with explicit supervision, helping the network to perceive 3D foreground points and further alleviate the feature-blurring problem. The segmentation scores are used to re-weight the voxel features to play the role of attention.

\item The detection head utilizes the fused BEV feature map to predict the 3D bounding box (center, size, and orientation) and category of the object in the scene.

\end{enumerate}

More details for each part are introduced in the following subsections. Note that the fusion method introduced below is based on the voxel-based implementation. A pillar-based version is provided at the end of this section.


\subsection{Image Branch}
The image branch is employed to extract multi-scale semantic features of the image, usually consisting of a backbone and a neck. The backbone extracts image features, and the neck fuses features with different receptive fields and sizes output by the backbone in different stages. The input of the image branch is an RGB image $\mathbf{I}\in \mathbb R^{H\times W \times 3}$, where $H$ and $W$ represent the height and width of the image, respectively. And the outputs are $n_I$ multi-scale feature maps $\mathbf{F}_{I,i}\in \mathbb R^{H_i\times W_i \times C_i}, i=1,2,\dots,n_I$, where $H_i$, $W_i$ and $C_i$ represent the height, width and channel dimensions of the $i$-th level, respectively. 
In our model, we adopt a classic configuration where the backbone is ResNet-50\cite{heDeepResidualLearning2016b} and the neck is FPN\cite{linFeaturePyramidNetworks2017a}.

\subsection{Voxel-Image Fusion Backbone}
Fig. \ref{fig:overall} shows a general voxel-image fusion backbone architecture, which consists of $n$ fusion blocks and $m$ ordinary blocks. For the input point cloud $\mathbf{P}\in \mathbb R^{N\times C_{in}}$, with $N$ points and $C_{in}$ channels, a sparse tensor $\mathcal{X}$ is obtained after voxelization, which can be expressed as a collection of 3D coordinates and non-empty voxel features, i.e., $\mathcal{X}=\{\mathbf{F}_V,\mathbf{C}_V\}$, where $\mathbf{F}_V\in \mathbb R^{N_V\times C_{in}}$  represents the features of $N_V$ non-empty voxels and $\mathbf{C}_V\in \mathbb R^{N_V\times 3}$ represents the coordinate of these voxels. 
After obtaining $\mathcal{X}$, it is passed through several fusion blocks and ordinary blocks. Following VoxelNeXt\cite{chenVoxelNeXtFullySparse2023a}, we employ 6 blocks, i.e., $n+m=6$.

\subsubsection{Ordinary Block}
An ordinary block is a stage in a commonly used sparse backbone such as SECOND\cite{yanSECONDSparselyEmbedded2018a} and VoxelNeXt\cite{chenVoxelNeXtFullySparse2023a}, which is generally composed of a sparse convolution layer used for downsampling and several submanifold convolution layers or residual blocks. The input of an ordinary block is a sparse tensor $\mathcal{X}_{in}$, and the output is another sparse tensor $\mathcal{X}_{out}$ with the same or downsampled spatial shape.
\subsubsection{General Fusion Block}
\begin{figure*}[!t]
    \centering
    \subfloat[][SFF]{\includegraphics[width=3.5in]{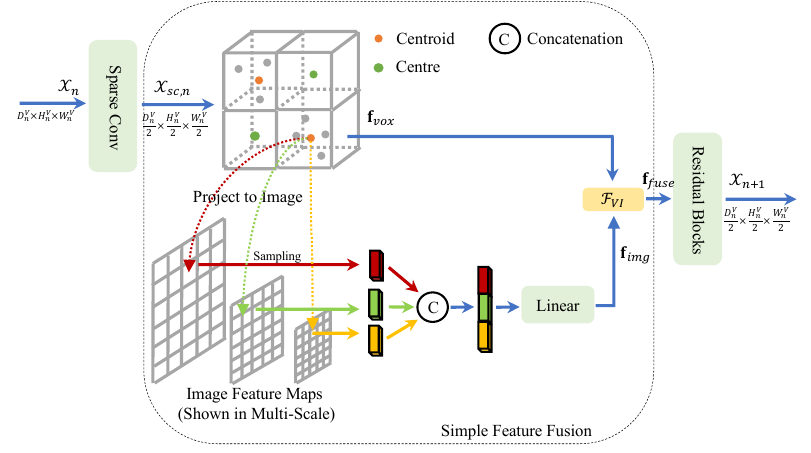}
    \label{fig:simple_fusion_block}}
    \hfil
    \subfloat[][MSDFF]{\includegraphics[width=3.5in]{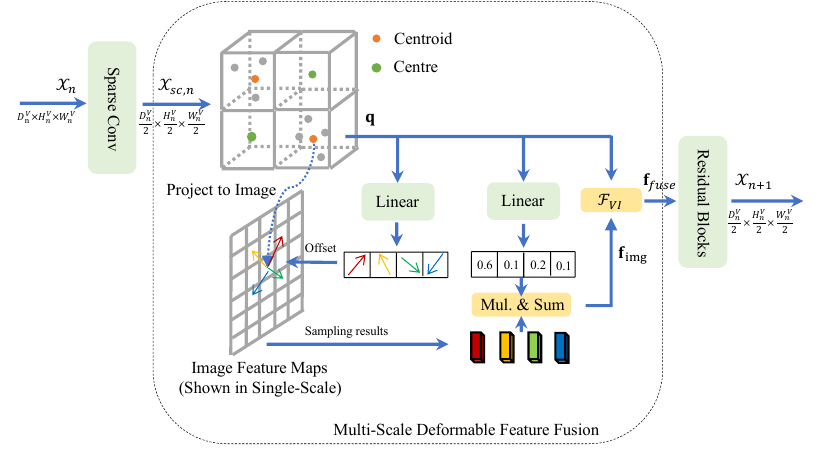}%
    \label{fig:deform_fusion_block}}
    \caption{Two types of fusion blocks. 
    (a) shows the fusion block based on the SFF. The sparse tensor $\mathcal{X}_{n}$ with spatial shape $(D^{V}_{n}, H^{V}_{n}, W^{V}_{n})$ output by the previous block is fed to a sparse convolution layer with stride 2. The centroid of each non-empty voxel (use the center instead if the centroid is not available) is projected onto the image. Image features are then sampled from the multi-scale image feature maps through bilinear interpolation. After concatenation and mapping, the sampled feature $\mathbf{f}_{img}$ is fused with the voxel feature $\mathbf{f}_{vox}$ through $\mathcal{F}_{VI}$, obtaining $\mathbf{f}_{fuse}$. The final output is obtained after several residual blocks.
    (b) shows the fusion block based on the MSDFF. Unlike (a), for a non-empty voxel, the corresponding query $\mathbf{q}$ is first passed through two parallel linear layers to obtain sampling offsets and weights. Image features are sampled from the multi-scale image feature maps according to the sampling offsets. After the weighted summation and fusion operator, the fused feature $\mathbf{f}_{fuse}$ is obtained. Other processes are consistent with (a).}
    \label{fig:fusion_block}
\end{figure*}

As shown in Fig. \ref{fig:fusion_block}, the fusion block can be regarded as an extension of the ordinary block. It provides an extra operation between the sparse convolution layer and the residual blocks to retrieve multi-scale image features provided by the image branch. Specifically, we first compute the centroids of non-empty voxels in the sparse tensor $\mathcal{X}_{sc}$ output by the sparse convolution. That is, for the $k$-th non-empty voxel $\mathcal{V}_k$ with feature $\mathbf{f}_{vox,k}$, assuming $\mathcal{P}_k=\{\mathbf{p}_j\}, \mathbf{p}_j\in \mathbb R^3, j=1,\cdots,|\mathcal{P}_k|$ is the points located in $\mathcal{V}_k$,
the centroid $\mathbf{c}_k \in \mathbb R^3$ can be obtained by
\begin{equation}
    \mathbf{c}_k=\frac{1}{|\mathcal{P}_k|}\sum_{j=1}^{|\mathcal{P}_k|}\mathbf{p}_j,
\end{equation}
where $|\cdot|$ represents taking the cardinality of a set. Note that ``non-empty voxel'' refers to a voxel with non-zero features in the sparse tensor, and its inferior may not necessarily contain radar points which is due to the dilation effect of sparse convolution\cite{chenFocalSparseConvolutional2022}. Thus, for non-empty voxel $\mathcal{V}_k$ with no internal points, i.e., $|\mathcal{P}_k|=0$, we use its center instead. For convenience, we refer to them as centroid as well. We omit the subscript $k$ in the following, which defaults to a single non-empty voxel. 

The centroid of each non-empty voxel is then projected onto the image according to the camera intrinsic matrix $\mathbf{T}_{intr} \in \mathbb R^{3\times4}$ and the radar-to-camera coordinate transformation matrix $\mathbf{T}_{r2c} \in \mathbb R^{4\times4}$, which can be formulated as
\begin{equation}
    \mathbf{c}_{img}' = \mathbf{T}_{intr} \cdot \mathbf{T}_{r2c} \cdot \mathbf{c}',
\end{equation}
where $\mathbf{c}'=[\mathbf{c}; 1]\in \mathbb R^4$ represents the homogeneous coordinates of $\mathbf{c}$. As a result, $\mathbf{c}_{img}'=[ud,vd,d]^T$ is the homogeneous coordinate in the image coordinate system where $(u,v)$ and $d$ denote the pixel index and the corresponding depth, respectively. Given the pixel index $\mathbf{c}_{img}=[u,v]^T$, the normalized coordinate can be obtained by $\tilde{\mathbf{c}}_{img} = [u/W, v/H]^T\in \mathbb R^{[0,1]\times[0,1]}$.

After obtaining the projection points for each non-empty voxel, image features are then retrieved from the multi-scale image feature maps by operator $\mathcal{E}$. Generally, the inputs of $\mathcal{E}$ are the normalized coordinate $\tilde{\mathbf{c}}_{img}$, multi-scale image features $\{\mathbf{F}_{I,i}\}_{i=1}^{n_I}$, and other useful input like queries denoted as $\mathbf{u}\in \mathbb R^{C_u}$. The output is the corresponding image feature $\mathbf{f}_{img}\in \mathbb R^{C_{img}}$, i.e., 
\begin{equation}
    \mathbf{f}_{img} = \mathcal{E}(\tilde{\mathbf{c}}_{img}, \{\mathbf{F}_{I,i}\}_{i=1}^{n_I}, \mathbf{u}).
\end{equation}
The image feature is then fused with the corresponding voxel feature $\mathbf{f}_{vox}$ by operator $\mathcal{F}_{VI}$, obtaining the fused feature $\mathbf{f}_{fuse}\in \mathbb R^{C_{fuse}}$, that is
\begin{equation}
    \mathbf{f}_{fuse}=\mathcal{F}_{VI}(\mathbf{f}_{img}, \mathbf{f}_{vox}).
\end{equation}
The fused sparse tensor $\mathcal{X}_{fuse}$ containing both voxel and image information can be constructed by simply replacing $\mathbf{f}_{vox}$ to $\mathbf{f}_{fuse}$ in $\mathcal{X}_{sc}$. Finally, $\mathcal{X}_{fuse}$ is passed through the residual block to further process and fuse neighborhood features, obtaining the final output $\mathcal{X}_{out}$. In our model, we let $C_{img}=C_{vox}=C_{fuse}$. Thus, the operators $\mathcal{E}$ and $\mathcal{F}_{VI}$ do not change the feature dimension of $\mathcal{X}_{sc}$. 

The above is a general strategy and no specific implementations of $\mathcal{E}$ and $\mathcal{F}_{VI}$ are given. In this study, we propose two implementations of the operator $\mathcal{E}$, i.e., simple feature fusion (SFF) and multi-scale deformable feature fusion (MSDFF). For $\mathcal{F}_{VI}$, we simply adopt addition, i.e.,
\begin{equation}
\label{equ:fusion_operator}
    \mathbf{f}_{fuse}=\mathcal{F}_{VI}(\mathbf{f}_{img}, \mathbf{f}_{vox})=\mathbf{f}_{img}+\mathbf{f}_{vox}.
\end{equation}

\subsubsection{Fusion Block Based on the SFF} 
The detailed process of the SFF is illustrated in Fig. \ref{fig:fusion_block}(a). According to the normalized image coordinate $\tilde{\mathbf{c}}_{img}$ projected from the centroid $\mathbf{c}$, we can use bilinear interpolation to sample image features from the multi-scale image feature maps. For the $i$-th level, we can obtain the sampled feature $\mathbf{f}_{img}^i$ by
\begin{equation}
\label{equ:simple_sample}
    \mathbf{f}_{img}^i={\tt{Sample}}(\mathbf{F}_{I,i}, \tilde{\mathbf{c}}_{img}),
\end{equation}
which can be accomplished using the ``$\tt{grid\_sample}$'' operation in Pytorch. The final image feature can be obtained by concatenating image features from all levels, and applying a linear project with batch normalization. The SFF operation is defined as:
\begin{equation}
\begin{aligned}
    \mathbf{f}_{img} &= \mathcal{E}_{SFF}(\tilde{\mathbf{c}}_{img}, \{\mathbf{F}_{I,i}\}_{i=1}^{n_I}) \\
            &= {\tt{Linear}_{BN}} ({\tt{Cat}}(\{\mathbf{f}_{img}^i\}_{i=1}^{n_I})).
\end{aligned}
\end{equation}

\subsubsection{Fusion Block Based on the MSDFF}
Simple feature fusion can only focus on a single location for each projected point, resulting in a limited receptive field. To utilize the surrounding information of objects, increase the receptive field, and further enhance the ability of the feature extraction operator $\mathcal{E}$, we propose multi-scale deformable feature fusion, i.e., MSDFF, as shown in Fig. \ref{fig:fusion_block}(b). 

The MSDFF is based on multi-scale deformable cross-attention\cite{zhuDeformableDETRDeformable2021a} which can select different sample positions and adjust the weight of sampled features according to the corresponding query. The process can be divided into two steps: query generation and deformable cross-attention.


\paragraph{Query Generation}
Extra queries are needed to guide the sampling process, and the queries are expected to have rich image and point cloud information to better select the desired features on the image feature maps. 

To meet this requirement, we add a query initialization module before the first fusion block to get high-quality queries. Specifically, for each reference point, we initialize the query as follows
\begin{equation}
\label{equ:init_query}
    \mathbf{q} = \mathbf{q}_{init} = {\tt{Linear_{BN}}}({\tt{Cat}}(\{\mathbf{f}_{vox}, \mathbf{f}_{img}^1, \mathbf{p}\})),
\end{equation}
where $\mathbf{p} \in \mathbb R^{N\times 3}$ 
is the normalized center coordinate of the corresponding voxel. As can be seen from Eq. (\ref{equ:init_query}), the initial query $\mathbf{q}_{init}$ combines information from three sources. Among them, the voxel feature $\mathbf{f}_{vox}$ provides radar information like RCS and velocity. The normalized center coordinate $\mathbf{p}$ indicates the position and distance of the corresponding voxel. As for $\mathbf{f}_{img}^1$, it is obtained by sampling the image feature in $\mathbf{F}_{I,1}$ following Eq. (\ref{equ:simple_sample}). As a result, $\mathbf{q}_{init}$ contains rich information from both modalities.

For the subsequent fusion block, we directly use $\mathbf{f}_{vox}$ as the query, since it already contains the previously fused image information, which means $\mathbf{q}=\mathbf{f}_{vox}$.

\paragraph{Deformable Cross-Attention}
Guided by the query, deformable cross-attention is employed for the sampling process. For clarity, only single-head attention is considered below. Suppose we have $n_I$ image feature maps and want to sample $n_s$ features from each feature map for one query. The query $\mathbf{q}$ is passed through two parallel linear layers to obtain offsets $\{\mathbf{o}_{i,j}=(o_{i,j}^x,o_{i,j}^y)\}, \mathbf{o}_{i,j}\in \mathbb R^{2}$ and weights $\{w_{i,j}\}, w_{i,j}\in \mathbb R$, where $i=1,\cdots, n_I$ and $j=1,\cdots,n_{s}$. The offsets are then normalized by the width and height of the corresponding image feature map, obtaining $\{\tilde{\mathbf{o}}_{i,j}=(o_{i,j}^x/W_i,o_{i,j}^y/H_i)\}$. For the $i$-th feature map, the $j$-th normalized sampling position $\tilde{\mathbf{c}}_{i,j}^s$ is calculated according to the reference point $\tilde{\mathbf{c}}_{img}$ and the offset $\tilde{\mathbf{o}}_{i,j}$ by
\begin{equation}
    \tilde{\mathbf{c}}_{i,j}^s = \tilde{\mathbf{c}}_{img} + \tilde{\mathbf{o}}_{i,j}.
\end{equation}

Then, we sample features from the corresponding image feature level by bilinear interpolation following Eq. (\ref{equ:simple_sample}) and weighted summation is performed according to the normalized weight $\{\tilde{w}_{i,j}\}={\tt Softmax} (\{w_{i,j}\})$. After applying a linear project with batch normalization, the image feature $\mathbf{f}_{img}$ is obtained. To sum up, the MSDFF operation can be formulated as:
\begin{equation}
\begin{aligned}
    \mathbf{f}_{img} &= \mathcal{E}_{MSDFF}(\tilde{\mathbf{c}}_{img}, \{\mathbf{F}_{I,i}\}_{i=1}^{n_I}, \mathbf{q}) \\
    &= {\tt{Linear_{BN}}} (\sum_{i=1}^{n_I}\sum_{j=1}^{n_{s}}\tilde{w}_{i,j}\cdot {\tt{Sample}}(\mathbf{F}_{I,i}, \tilde{\mathbf{c}}_{i,j}^s)).
\end{aligned}
\end{equation}

Through the above method, the point cloud features are deeply interacted with the image features, so that the fused features contain rich geometric and semantic information, provided by the point clouds and images, respectively. These features help the network distinguish 3D foreground points and alleviate the feature-blurring problem.




\subsection{Semantic-Guided Head}
To better utilize the fused features and further mitigate the feature-blurring problem, we apply the semantic-guided head to the output of the last fusion block.

In particular, for each non-empty voxel, we pass its feature $\mathbf{f}_{fuse}$ through a multi-layer perceptron (MLP) and obtain the corresponding foreground score $s_{seg}$ to predict whether this voxel is a foreground one, i.e.,
\begin{equation}
    s_{seg} = {\tt{Sigmoid}}({\tt{MLP}}(\mathbf{f}_{fuse})).
\end{equation}
A foreground voxel is defined as a non-empty voxel whose centroid is in a 3D ground truth bounding box. Otherwise, we define it as a background voxel. Focal Loss\cite{linFocalLossDense2017} is used for explicit supervision. When the score of each non-empty voxel is obtained, the voxel features are further multiplied by the score to guide the network to pay more attention to 3D foreground points. The weighted feature is obtained by
\begin{equation}
    \mathbf{f}_{fuse}' = \mathbf{f}_{fuse}\cdot s_{seg}.
\end{equation}

\subsection{3D Neck \& Detection Head}
It should be noted that we mainly focus on the voxel-image fusion backbone, especially the fusion strategy. The 3D neck and the detection head are not limited to specific methods. For our voxel version, we adopt the methodology of VoxelNeXt\cite{chenVoxelNeXtFullySparse2023a}. The following is a brief introduction. Please refer to \cite{chenVoxelNeXtFullySparse2023a} for more details. 

Given the sparse tensor $\mathcal{X}_{out,i}=\{\mathbf{F}_{V,i},\mathbf{C}_{V,i}\}$ output by the block $i$, the 3D neck combines the non-empty voxels contained in $\{\mathcal{X}_{out,4}, \mathcal{X}_{out,5}, \mathcal{X}_{out,6}\}$. Specifically, taking $\mathbf{C}_{V,4}$ as a reference, the voxel coordinates $\mathbf{C}_{V,i}, i>4$ is multiplied by its downsampling stride relative to $\mathbf{C}_{V,4}$ to align the voxel coordinates, e.g., $\mathbf{C}_{V,5}'=\{2\times x,2\times y,2\times z | (x,y,z)\in \mathbf{C}_{V,5}\}$. A new sparse tensor $\mathcal{X}_{comb}$ can be constructed by $\mathcal{X}_{comb}=\{\cup_{i=4}^6 \mathbf{F}_{V,i},\cup_{i=4}^6 \mathbf{C}_{V,i}'\}$. All the features corresponding to the same X and Y coordinates are added to obtain the output BEV feature map.

The detection head of VoxelNeXt is a sparse version of center head\cite{yinCenterbased3DObject2021} and has lower computational costs\cite{chenVoxelNeXtFullySparse2023a}. 

\subsection{Extend to Pillar}

\begin{figure}[!t]
\centering
\includegraphics[width=3.0in]{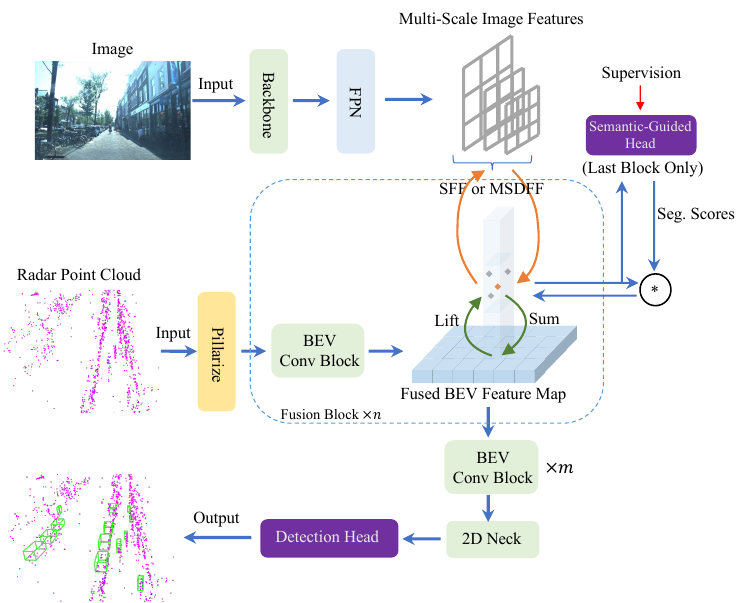}
\caption{The pillar version of the proposed method.}
\label{fig:pillar_version}
\end{figure}

    To facilitate the introduction of the proposed fusion module, the descriptions above are based on the voxel-based backbone. But our approach can be easily extended to lightweight yet powerful pillar-based networks with slightly adaptation. 

    Fig. \ref{fig:pillar_version} presents the pillar version of our method. The major difference from the voxel version is that, in the pillar version, the intermediate features in the backbone are 2D BEV features rather than 3D sparse tensors. This difference does not affect the application of our fusion methodology. Specifically, for pillar-based networks, the input point cloud is typically pillarized before being fed into a BEV backbone with several conventional blocks. We replace the first $n$ blocks with our fusion block, as shown in Fig. \ref{fig:pillar_version}.
    For the BEV feature output by a conventional block, the original point cloud can be voxelized with the same X and Y-resolution. For any non-empty voxel $\mathcal{V}$ with coordinates $(x,y,z)$, we can index the feature from the corresponding BEV feature map based on $(x,y)$. This feature can be directly lifted as the voxel's feature. However, to preserve height information, we introduce a learnable height embedding. The final voxel feature $\mathbf{f}_{vox}$ is obtained by adding the corresponding BEV feature and the height embedding, expressed as
    \begin{equation}
        \mathbf{f}_{vox} = {\tt{BEV}}(x, y) + {\tt{HeightEmb}}(z),
    \end{equation}
    where ${\tt{BEV}}(x, y)$ represents the BEV feature at $(x, y)$, and ${\tt{HeightEmb}}(z)$ represents the height embedding corresponding to height $z$.
    In this way, the BEV feature is effectively lifted into the non-empty voxels within 3D space. The SFF and MSDFF described above can then be used to aggregate features from images and fuse them into the non-empty voxels. Then, the voxel features with identical $(x,y)$ coordinates are summed 
    to produce the fused BEV feature. Similar to the voxel-based version, we also employ a semantic-guided head after the last fusion block. Finally, the fused BEV feature pass through the remaining convolutional blocks, followed by the 2D neck and detection head, to produce the final detection results.



\subsection{Loss}
Compared with the loss of the single-modal 3D object detection network, our method includes an additional segmentation loss in the semantic-guided head. Assuming $L_{seg}$ is the segmentation loss based on Focal Loss\cite{linFocalLossDense2017}, and $L_{det}$ is the detection loss corresponding to the single-modal method\cite{chenVoxelNeXtFullySparse2023a}\cite{langPointPillarsFastEncoders2019a}, which is generally composed of classification, location and other losses depending on the specific method. The total loss can be obtained by
\begin{equation}
    L = \alpha_1L_{seg} + \alpha_2L_{det},
\end{equation}
where $\alpha_1$ and $\alpha_2$ are the weights used to balance these two losses. In our model, we simply set $\alpha_1=\alpha_2=1$.



\section{Experments and analysis}
\label{sec:exps}
\subsection{Dataset and Evaluation Metrics}
\subsubsection{Dataset}
We use two datasets to evaluate our model, i.e., the VoD dataset\cite{palffyMultiClassRoadUser2022} and the TJ4DRadset dataset\cite{zhengTJ4DRadSet4DRadar2022}. Both of them are oriented toward autonomous driving applications, especially for 4D radar perception.

The VoD dataset has a total of 8682 frames, including 5139 frames in the training set, 1296 frames in the validation set, and 2247 frames in the test set. Since the official evaluation system is not yet open, our comparison and ablation experiments were completed on the validation set. The VoD dataset is collected in the city of Delft (The Netherlands) and covers campus, suburb and old-town scenarios. It provides synchronized 4D radar, LiDAR, camera, and GPU/IMU data with 3D annotations and tracking IDs. Moreover, the official also provides radar point clouds accumulated from multiple scans, which is implemented by compensating ego-motion. Following previous works\cite{liuSMURFSpatialMultiRepresentation2023a}\cite{xiongLXLLiDARExcluded2023b}\cite{zhengRCFusionFusing4D2023}, we use the five-scan radar points and consider three categories, i.e., car, pedestrian, and cyclist.

The TJ4DRadset dataset was collected in Suzhou, China, covering different road types, such as urban roads, elevated roads and industrial zones. Compared with the VoD dataset, it contains more difficult scenarios, e.g., nighttime, glare, under the bridge, and wrong camera focus, which are big challenges for the camera. It has a total of 7746 frames, of which 5706 frames are for training and 2040 frames for testing. Its sensor configuration is similar to that in the VoD dataset, but it only provides synchronized 4D radar and camera data, with 3D annotations and tracking IDs up to now. For the TJ4DRadset dataset, we adopt single-frame radar data without accumulation and consider four categories, i.e., car, pedestrian, cyclist, and truck.

\subsubsection{Evaluation Metrics}
Both the VoD and TJ4DRadSet datasets use averaging precision (AP) as the main evaluation metric. 

For the VoD dataset, according to the official recommendations, two metrics are used, i.e., AP under the entire annotated area ($\text{AP}_\text{EAA}$) and AP under the driving corridor ($\text{AP}_\text{DC}$). The former means that all annotations are used for evaluation regardless of range. The latter means that we only consider annotations located in a specific area which is defined as $\mathcal{A}_{DC}=\{(x,y,z)|-4\text{m}<x<4\text{m}, z<25\text{m}\}$ in camera coordinates. When calculating AP, the intersection-over-union (IoU) threshold is set to 0.25 for cyclists and pedestrians, and 0.5 for cars. The IoU threshold is used to determine positive and negative samples. 

For the TJ4DRadset dataset, 3D AP ($\text{AP}_\text{3D}$) and BEV AP ($\text{AP}_\text{BEV}$) are evaluated with a uniform range of 0-70m, and the IoU thresholds are set in line with the VoD dataset, with an IoU threshold of 0.5 for the additional truck category.

\subsection{Implementation Details}
We implement our model based on MMDetection3D\cite{mmdet3d2020} framework which is an open-source 3D object detection toolbox based on PyTorch. For the VoD dataset, following official configurations, we set the point cloud range to $\{(x,y,z)|0\text{m}<x<51.2\text{m},-25.6\text{m}<y<25.6\text{m},-3\text{m}<z<2\text{m}\}$. We use the radar point cloud accumulated over 5 scans as input. The input feature is selected as
\begin{equation}
    \mathbf{f}_{in}^{\text{VoD}} = [x, y, z, RCS, v_r, v_{rc}, t]^T,
\end{equation}
where $RCS$ represents the radar cross section reflecting the reflection intensity of the object, $v_r$ is the relative radial Doppler velocity, $v_{rc}$ is the absolute radial Doppler velocity, and $t$ is the time ID, indicating which scan it originates from.

For the TJ4DRadset dataset, as previous works\cite{yinCenterbased3DObject2021}\cite{chenVoxelNeXtFullySparse2023a} on KITTI\cite{geigerAreWeReady2012}, we use slightly different point cloud ranges for the voxel-base and pillar-based methods, to make the point cloud ranges divisible by the size of the voxel or pillar. Specifically, for pillar-based methods, we set the point cloud range to $\{(x,y,z)|0\text{m}<x<69.12\text{m},-39.68\text{m}<y<39.68\text{m},-4\text{m}<z<2\text{m}\}$. For voxel-based methods, we set the point cloud range to $\{(x,y,z)|0\text{m}<x<70.4\text{m},-40\text{m}<y<40\text{m},-4\text{m}<z<2\text{m}\}$. The input feature is selected as
\begin{equation}
    \mathbf{f}_{in}^{\text{TJ4D}} = [x, y, z, v_{rc}, Power]^T,
\end{equation}
where $v_{rc}$ has the same definition as that in the VoD dataset, $Power$ is in dB scale and represents the signal-to-noise ratio of the detection. \textit{It should be noted that the official public data of the TJ4DRadset dataset only provide the relative radial Doppler velocity $v_r$ and no IMU data. Following \cite{zhengRCFusionFusing4D2023}, we use the method in \cite{tan3DObjectDetection2023} to estimate ego-motion to obtain the absolute radial Doppler velocity $v_{rc}$.}

In both datasets, the voxel sizes are set to 0.05m, 0.05m and 0.125m along the X-, Y- and Z-axis, respectively. The pillar sizes are set to 0.16m, 0.16m along the X- and Y-axis, respectively. We use the hybrid task cascade network (HTC)\cite{chenHybridTaskCascade2019} provided by MMDetection\cite{mmdetection} to initialize the image branch and freeze its parameters during training. The HTC is pre-trained on COCO\cite{linMicrosoftCOCOCommon2015a} and fine-tuned on nuImage\cite{caesarNuScenesMultimodalDataset2020a}. For PointPillars, predefined anchor boxes are needed. As official settings, for the VoD dataset, the dimensions of anchor boxes for the car, pedestrian, and cyclist categories are (3.9m, 1.6m, 1.56m), (0.8m, 0.6m, 1.73m), and (1.76m, 0.6m, 1.73m), respectively. For the TJ4DRadset dataset, the dimensions of anchor boxes for the car, pedestrian, cyclist, and truck categories are (1.84m, 4.56m, 1.70m), (0.6m, 0.8m, 1.69m), (0.78m, 1.77m, 1.60m), and (2.66m, 10.76m, 3.47m), respectively. For VoxelNeXt, compared with the original configuration, the voxel features in our backbone need to contain both point cloud and image information, so we double the output feature dimensions of all blocks. By default, we adopt the MSDFF-based fusion block and set $n=2$. Other configurations show equally good performance, which is confirmed in the ablation experiment section.

The network is trained on a single NVIDIA RTX 3090 graphic processing unit (GPU) by the AdamW optimizer and one-cycle learning rate scheduler in an end-to-end manner. The initial learning rate is set to 0.001. We use random flip, random scaling and random rotation data augmentation for the input point cloud, and no data augmentation is used for the input image. 

\subsection{Experiment Results}

\begin{table*}[!t]
\caption{Detection Results on VoD\label{tab:vod_results}}
\centering
\resizebox{\linewidth}{!}{
\begin{tabular}{c|c|ccc|c|ccc|c|c}
\hline \multirow{2}{*}{ Method } & \multirow{2}{*}{ Modality } & \multicolumn{4}{c|}{ AP in the Entire Annotated Area (\%) } & \multicolumn{4}{c|}{AP in the Driving Corridor (\%) } & \multirow{2}{*}{FPS} \\
\cline { 3 - 10 } & & Car & Pedestrian & Cyclist & mAP & Car & Pedestrian & Cyclist & mAP \\
\hline 
ImVoxelNet (WACV 2022)\cite{Rukhovich2022ImVoxelNet} 
&$\mathrm{C}$
& 19.35 & 5.62 & 17.53 & 14.17 & 49.52 & 9.68 & 28.97 & 29.39 & 11.1 \\
\hline
PointPillars (CVPR 2019)\cite{langPointPillarsFastEncoders2019a} 
&$\mathrm{R}$
& 42.19 & 39.29 & 66.66 & 49.38 & 71.59 & 50.67 & 85.23 & 69.16 & \textbf{106.4} \\
RadarPillarNet (IEEE T-IM 2023)\cite{zhengRCFusionFusing4D2023} 
&$\mathrm{R}$
& 39.30 & 35.10 & 63.63 & 46.01 & 71.65 & 42.80 & 83.14 & 65.86 & - \\
CenterPoint (CVPR 2022)\cite{yinCenterbased3DObject2021}  
&$\mathrm{R}$
& 35.84 & 41.03 & 67.11 & 47.99 & 70.65 & 50.14 & 85.67 & 68.82 & 38.3 \\
VoxelNeXt (CVPR 2023)\cite{chenVoxelNeXtFullySparse2023a}        
&$\mathrm{R}$
& 36.98 & 42.37 & 68.15 & 49.17 & 70.95 & 51.85 & 87.33 & 70.04 & 31.6 \\
SMURF (IEEE T-IV 2023)\cite{liuSMURFSpatialMultiRepresentation2023a}        
&$\mathrm{R}$
& 43.31 & 39.09 & 71.50 & 50.97 & 71.74 & 50.54 & 86.87 & 69.72 & - \\
MSSF-V-R        
&$\mathrm{R}$
& 38.28 & 42.93 & 69.96 & 50.39 & 71.76 & 52.92 & 88.93 & 71.21 & 24.6 \\
MSSF-PP-R        
&$\mathrm{R}$
& 42.17 & 40.28 & 65.41 & 49.29 & 72.04 & 51.06 & 83.09 & 68.73 & 104.9 \\
\hline
PointAugmenting (CVPR 2021)\cite{wangPointAugmentingCrossModalAugmentation2021} 
&$\mathrm{R}+\mathrm{C}$
& 39.62 & 44.48 & 73.70 & 52.60 & 71.02 & 48.59 & 87.57 & 69.06 & $7.9^\dag$ \\

$\text{Focals Conv}^\ddag$ (CVPR 2022)\cite{chenFocalSparseConvolutional2022} 
&$\mathrm{R}+\mathrm{C}$
& 40.01 & 48.67 & 75.42 & 54.70 & 71.79 & 53.41 & 87.53 & 70.91 & $10.4^\dag$ \\

RCFusion (IEEE T-IM 2023)\cite{zhengRCFusionFusing4D2023} 
&$\mathrm{R}+\mathrm{C}$
& 41.70 & 38.95 & 68.31 & 49.65 & 71.87 & 47.50 & 88.33 & 69.23 & - \\

LXL (IEEE T-IV 2023)\cite{xiongLXLLiDARExcluded2023b}
&$\mathrm{R}+\mathrm{C}$
& 42.33 & 49.48 & 77.12 & 56.31 & 72.18 & 58.30 & 88.31 & 72.93 & $6.1^*$\\

UniBEVFusion (Arxiv 2024) \cite{zhao2024unibevfusion}
&$\mathrm{R}+\mathrm{C}$ & 42.22 & 47.11 & 72.94 & 54.09 & 72.10 & 57.71 & \textbf{93.29} & 74.37 & - \\

\hline




MSSF-V (\textbf{Ours}) 
&$\mathrm{R}+\mathrm{C}$ 
& 52.53 & \textbf{51.58} & 75.77 & 59.96 & 89.08 & \textbf{66.78} & 88.10 & \textbf{81.32} & 10.3\\

MSSF-PP (\textbf{Ours}) 
&$\mathrm{R}+\mathrm{C}$ 
& 60.96 & 51.28 & \textbf{77.69} & \textbf{63.31} & 90.60 & 60.39 & 88.35 & 79.78 & 13.9 \\

\hline
PointPillars (CVPR 2019)\cite{langPointPillarsFastEncoders2019a} 
&$\mathrm{L}$
& \textbf{68.81} & 51.26 & 66.00 & 62.02 & \textbf{90.84} & 62.80 & 85.25 & 79.63 & 56.1\\
\end{tabular}
}
\end{table*}

Since there are relatively few models designed specifically for 4D radar, we choose some methods originally designed for LiDAR for comparison. In Table \ref{tab:vod_results} and Table \ref{tab:tj4d_results}, PointPillars\cite{langPointPillarsFastEncoders2019a}, CenterPoint\cite{yinCenterbased3DObject2021}, and VoxelNeXt\cite{chenVoxelNeXtFullySparse2023a} are reproduced by us under the MMDetection3D framework. Focals Conv\cite{chenFocalSparseConvolutional2022} and PointAugmenting\cite{wangPointAugmentingCrossModalAugmentation2021} are reproduced according to the configuration in their official GitHub repositories with minimal changes to adapt to the 4D radar datasets. Focals Conv marked with $\ddag$ means using the same radar backbone as ours. Additionally, the detection results of PointPillars based on LiDAR and ImVoxelNet\cite{Rukhovich2022ImVoxelNet} based on monocular camera are also provided. For the VoD dataset, the inference speed in frames per second (FPS) is measured. The FPS results with $*$ and $\dag$ represent the use of different GPUs and different implementation frameworks, respectively.

For our method, multiple variants are given for comparison. Among them, MSSF-PP and MSSF-V represent the multi-modal versions of our method based on PointPillars and VoxelNeXt, respectively, while MSSF-PP-R and MSSF-V-R are the corresponding single-modal networks.


\subsubsection{Results on VoD Dataset}
In Table \ref{tab:vod_results}, we report the results on the validation set of the VoD dataset. 

\paragraph{The Characteristic of the Metrics} It can be seen that the detection results of the cyclist class are consistently good for all models, even outperforming the detection capabilities of LiDAR. This is because most of the cyclists in the dataset are in motion\cite{palffyMultiClassRoadUser2022}, and radar can measure the radial Doppler velocity of the object. Hence, radar is more sensitive to moving objects and can achieve good detection results even in a single modal. In contrast, there are many stationary objects in cars and pedestrians, which are not friendly to radar and easily confused with noisy background points, leading to poor detection results. Moreover, the difference between $\text{AP}_{\text{EAA}}$ and $\text{AP}_{\text{DC}}$ reflects that radar has better detection performance for close objects, as ones contain more detection points. In addition, the detection results of ImVoxelNet\cite{Rukhovich2022ImVoxelNet} indicate that despite high resolution and rich semantic information provided by camera, the lack of depth information leads to poor performance, especially for distant objects, as evidenced by $\text{mAP}_{\text{DC}}$ being greater than $\text{mAP}_{\text{EAA}}$.

\paragraph{Comparison with the State-of-the-Art Methods} The experiment results also show that our methods outperform the others in almost all metrics.  
Comparing MSSF-PP with the latest strong published benchmark LXL\cite{xiongLXLLiDARExcluded2023b}, we achieved significant improvements of 7.0\% and 6.9\% on $\text{mAP}_\text{EAA}$ and $\text{mAP}_\text{DC}$, respectively. For MSSF-V, there are also 3.7\% and 8.4\% improvements. The consistent improvements shown by MSSF-PP and MSSF-V illustrate the versatility of our method. Specifically, for the car category, MSSF-PP and MSSF-V outperform LXL by 18.6\% and 10.2\% under $\text{AP}_{\text{EAA}}$, respectively. $\text{AP}_{\text{DC}}$ also has a notably substantial improvement of 18.4\% and 16.9\%, respectively, which is much higher than other categories. The reason is that the number of radar points on cars is relatively large, so there are more reference points projected onto the image, and sufficient semantic information can be captured. 
For the pedestrian category, there are also improvements of 1.8\% and 2.1\% respectively on $\text{AP}_{\text{EAA}}$, which shows the effectiveness of the proposed method. It is worth noting that our method does not improve significantly on the cyclist category, which we attribute to the fact that cyclists are already easier to detect even in single-modal, as mentioned in the above analysis. During the experiment, we also find that cyclists and pedestrians are occasionally confused after fusing image features. This is because bicycles are sometimes confused with the background, and the riders above are likely mistaken for pedestrians.


\begin{figure*}[!t]
    \centering
    \includegraphics[width=7in]{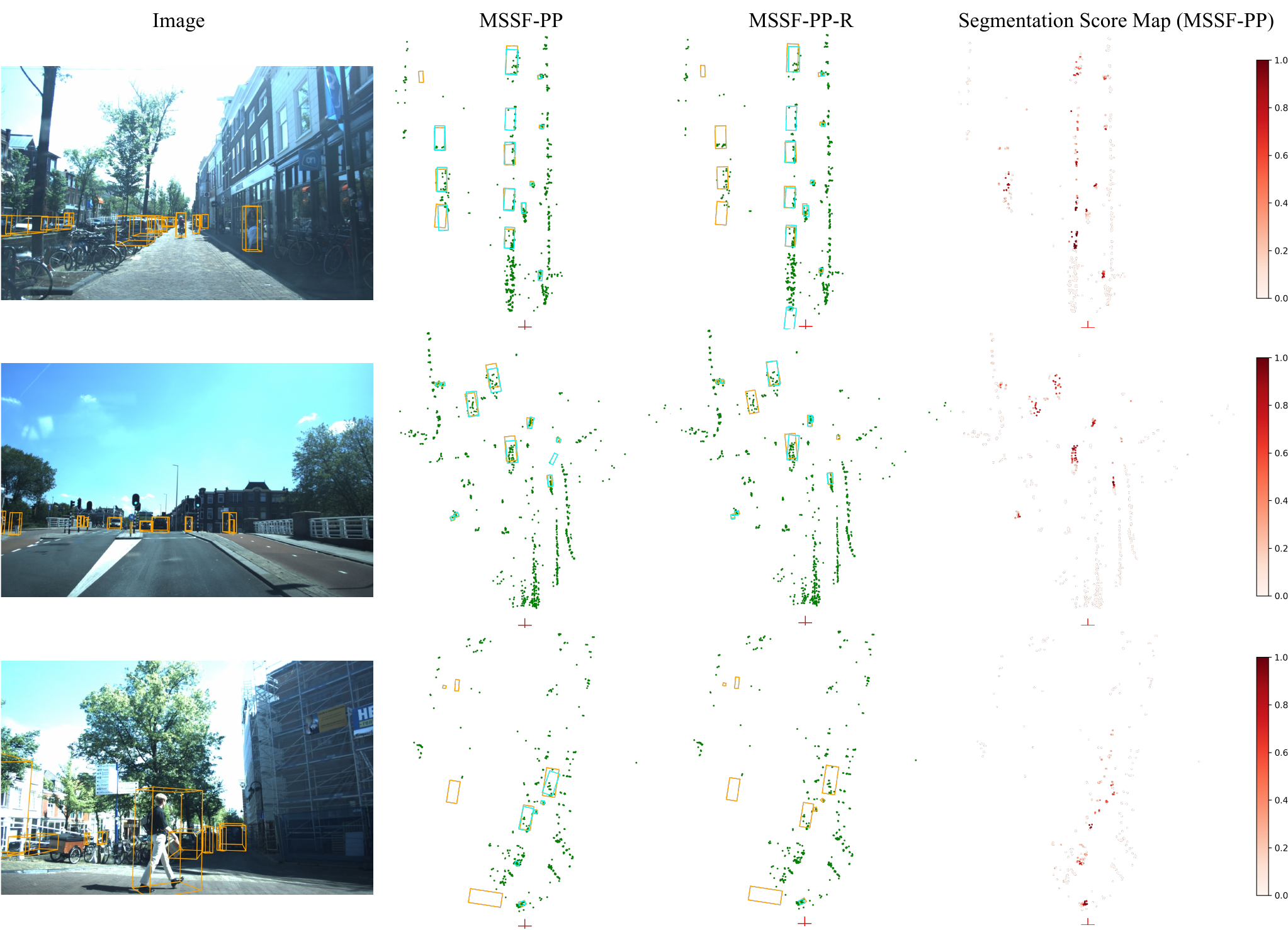}
    \caption{Visualization results on the VoD dataset (best viewed in color and zoom). Each row represents a frame. The first column shows the image, where the orange boxes represent ground truth. The second column shows the detection results of MSSF-PP from the BEV perspective, where the green points are radar points, the red crosses represent the self-vehicle position, the orange boxes represent ground truth bounding boxes and the cyan boxes represent predicted bounding boxes. The third column is the detection results of the single-modal version MSSF-PP-R with the same meaning as the second column. The fourth column shows the visualization results of the segmentation scores output by the semantic-guided head under BEV (darker colors indicate higher scores).}
    \label{fig:vis_vod}
\end{figure*}

\paragraph{Comparison with the Methods Designed for LiDAR-Camera Fusion} Our methods also demonstrate advantages when compared with Focals Conv\cite{chenFocalSparseConvolutional2022} and PointAugmenting\cite{wangPointAugmentingCrossModalAugmentation2021} which are proposed in the context of the fusion of LiDAR and camera. 
As for Focals Conv, MSSF-PP performs better in all three categories, improving 21.0\%, 2.6\%, and 2.3\% under $\text{AP}_\text{EAA}$ for cars, pedestrians, and cyclists, respectively.
Although Focals Conv explicitly classifies foreground and background points which is similar to us, it mainly focuses on its focal sparse convolution and only uses a shallow network as the image backbone. The absorbed image features have insufficient semantic expression capabilities. For PointAugmenting, it uses DLA-34\cite{yuDeepLayerAggregation2018} as the image backbone with richer image features, but it only fuses image features in the early stage without additional segmentation. Our method still demonstrates better performance than PointAugmenting. The above experiments reveal that the proposed fusion blocks and fusion strategy can effectively absorb and utilize image features.

\paragraph{Comparison with PointPillars under LiDAR Modality} 
We narrow the performance gap with LiDAR-based methods. 
Compared to the classic PointPillars\cite{langPointPillarsFastEncoders2019a} model in the LiDAR modality, we achieve superior performance both within the entire annotated area (EAA) and driving corridor (DC), with significantly lower costs. Specifically, we notably outperforms PointPillars by 11.7\% under $\text{AP}_\text{EAA}$ for the cyclist category.
Although PointPillars exhibits a considerable advantage in detecting cars, it is important to note that this is partly attributed to the distinct installation positions of LiDAR (on the roof) and 4D radar (behind the front bumper), which results in a limited field of view of radar, especially when considering the EAA.


\paragraph{Inference Speed} As for the inference speed, MSSF-PP and MSSF-V achieved 13.9 and 10.3 FPS, respectively. A quasi-real-time detection speed is achieved with significantly better performance than other methods, without dedicated code optimization.

\paragraph{Visualization} Fig. \ref{fig:vis_vod} shows the visualization results of MSSF-PP and MSSF-PP-R on the VoD dataset, indicating that the proposed fusion strategy can make good use of two modalities. Stationary and distant objects are weaknesses of single-modal models. Our method can use image information to reduce missed detections, e.g., for the cars in the opposite lane on the left side shown in the first row. In addition, when the object is far away or occluded (e.g., pedestrians 
at a distance shown in the first and second rows), our model can still use radar information for detection, showing certain modal robustness. The fourth column demonstrates that our model can distinguish 3D foreground points well, thus guiding the network to focus on foreground objects without triggering excessive false alarms caused by the feature-blurring problem.


\begin{table*}[!t]
\caption{Detection Results on TJ4DRadSet\label{tab:tj4d_results}}
\centering
\begin{tabular}{c|c|cccc|c|cccc|c}
\hline \multirow{2}{*}{ Method } & \multirow{2}{*}{ Modality } & \multicolumn{5}{c|}{ $\text{AP}_{\text{3D}}$ (\%) } & \multicolumn{5}{c}{ $\text{AP}_{\text{BEV}}$ (\%) } \\
\cline { 3 - 12 } & & Car & Ped. & Cyclist & Truck & mAP & Car & Ped. & Cyclist & Truck & mAP \\
\hline
ImVoxelNet (WACV 2022)\cite{Rukhovich2022ImVoxelNet} 
&$\mathrm{C}$
& 15.69 & 9.63 & 16.66 & 13.53 & 13.87 & 22.35 & 11.18 & 17.12 & 18.17 & 17.21 \\
\hline 
PointPillars (CVPR 2019)\cite{langPointPillarsFastEncoders2019a} 
&$\mathrm{R}$
& 19.78 & 29.79 & 51.83 & 13.67 & 28.76 & 39.42 & 32.56 & 59.45 & 20.93 & 38.09 \\
RadarPillarNet (IEEE T-IM 2023)\cite{zhengRCFusionFusing4D2023} 
&$\mathrm{R}$
& 28.45 & 26.24 & 51.57 & 15.20 & 30.37 & 45.72 & 29.19 & 56.89 & 25.17 & 39.24 \\
CenterPoint (CVPR 2022)\cite{yinCenterbased3DObject2021}  
&$\mathrm{R}$
& 11.23 & 25.47 & \textbf{56.20} & 4.95 & 24.47 & 24.01 & 29.46 & \textbf{61.08} & 8.03 & 30.65  \\
VoxelNeXt (CVPR 2023)\cite{chenVoxelNeXtFullySparse2023a}        
&$\mathrm{R}$
& 13.27 & 33.54 & 52.59 & 8.32 & 26.93 & 23.17 & 35.83 & 57.11 & 12.12 & 32.06  \\
SMURF (IEEE T-IV 2023)\cite{liuSMURFSpatialMultiRepresentation2023a}        
&$\mathrm{R}$
& 28.47 & 26.22 & 54.61 & 22.64 & 32.99 & 43.13 & 29.19 & 58.81 & 32.80 & 40.98 \\
MSSF-V-R        
&$\mathrm{R}$
& 12.34 & 31.73 & 53.16 & 9.15 & 26.60 & 22.85 & 33.03 & 58.69 & 14.70 & 32.32 \\
MSSF-PP-R        
&$\mathrm{R}$
& 21.08 & 31.99 & 50.39 & 10.36 & 28.45 & 38.47 & 35.97 & 57.98 & 22.01 & 38.61 \\
\hline
PointAugmenting (CVPR 2021)\cite{wangPointAugmentingCrossModalAugmentation2021} 
&$\mathrm{R}+\mathrm{C}$
& 22.63 & 26.23 & 53.52 & 13.37 & 28.94 & 43.42 & 29.65 & 59.21 & 23.88 & 39.04 \\
$\text{Focals Conv}^\ddag$ (CVPR 2022)\cite{chenFocalSparseConvolutional2022} 
&$\mathrm{R}+\mathrm{C}$
& 12.24 & 31.80 & 54.01 & 6.66 & 26.18 & 22.52 & 35.33 & 59.37 & 10.30 & 31.88 \\
RCFusion (IEEE T-IM 2023)\cite{zhengRCFusionFusing4D2023} 
&$\mathrm{R}+\mathrm{C}$
& 29.72 & 27.17 & 54.93 & 23.56 & 33.85 & 40.89 & 30.95 & 58.30 & 28.92 & 39.76\\
LXL (IEEE T-IV 2023)\cite{xiongLXLLiDARExcluded2023b}       
&$\mathrm{R}+\mathrm{C}$
&   -   &   -   &   -   &   -   & 36.32 &   -   &   -   &   -   &   -   & 41.20 \\

UniBEVFusion (Arxiv 2024) \cite{zhao2024unibevfusion}   
&$\mathrm{R}+\mathrm{C}$
& 44.26 & 27.92 & 51.11 & \textbf{27.75} & 37.76 & 50.43 & 29.57 & 56.48 & \textbf{35.22} & 42.92 \\
\hline

MSSF-V (\textbf{Ours}) 
&$\mathrm{R}+\mathrm{C}$
& 45.18 & 33.61 & 55.88 & 17.20 & 37.97 & 56.25 & 36.53 & 58.70 & 20.97 & 43.11  \\

MSSF-PP (\textbf{Ours})
&$\mathrm{R}+\mathrm{C}$
& \textbf{52.04} & \textbf{35.11} & 55.72 & 24.14 & \textbf{41.75} & \textbf{64.31} & \textbf{38.39} & 60.08 & 30.86 & \textbf{48.41}  \\

\hline
\end{tabular}
\end{table*}

\subsubsection{Results on TJ4DRadSet Dataset}
Compared with VoD, TJ4DRadset is a more challenging dataset because it contains complex scenes such as nighttime, under bridges, and camera out-of-focus. In these scenarios, the image quality degrades significantly. Object detection in these difficult scenarios requires good cooperation between different modalities. In addition, TJ4DRadset has an additional truck category, and the size of objects in this category varies greatly, further increasing the difficulty of detection. Similar to Table \ref{tab:vod_results}, we present the results of the single-modal baseline, multi-modal baseline, and our methods on the TJ4DRadset test set in Table \ref{tab:tj4d_results}.

\paragraph{The Characteristic of the Metrics} As can be seen from the results, the AP is obviously lower than that of the VoD dataset, indicating that the TJ4DRadSet is more challenging as mentioned above. Note that, the pillar-based method has more advantages in this dataset. The possible reason is that the 4D radar used in the TJ4DRadset dataset is different from that in VoD and there is no multi-frame accumulation. The resulting point cloud distribution is different, which makes it hard for the voxel-based method to effectively extract point cloud features. This observation is consistent with \cite{zhengRCFusionFusing4D2023}. In addition, for cars and trucks with relatively large and various dimensions, the anchor-free method is inferior to the anchor-based ones. Moreover, compared to ImVoxelNet\cite{Rukhovich2022ImVoxelNet}, radar-based methods demonstrate significant advantages in detecting moving objects (e.g., pedestrians and cyclists).

\paragraph{Comparison with Other Methods} Even if the image quality is poor in some scenes, our method still achieves significant improvements. Compared to the corresponding single-modal baselines, the MSSF-PP and MSSF-V exhibit improvements on $\text{mAP}_\text{3D}$ by 13.3\% and 11.4\%, respectively, demonstrating the effectiveness of our proposed fusion network. 
Furthermore, our MSSF-PP also outperforms the recent state-of-the-art method UniBEVFusion\cite{zhao2024unibevfusion} by 4.0\% and 5.5\% on $\text{mAP}_\text{3D}$ and $\text{mAP}_\text{BEV}$, respectively. 
In line with the results observed on the VoD dataset, our method exhibits the largest improvement in the car category. PointAugmenting\cite{wangPointAugmentingCrossModalAugmentation2021} and Focals Conv\cite{chenFocalSparseConvolutional2022} do not show the advantages of multi-modal methods due to the use of weak image features and simple fusion strategies. These results indicate that our approach can effectively leverage multimodal features and achieve better performance, even in challenging and complex scenarios.

\paragraph{The Impact of Lighting Conditions} To analyze the impact of image quality on multi-modal methods in more detail, following LXL, we classify the sequences in the TJ4DRadset dataset according to different image degradations. Specifically, we classify the sequence into dark, dazzle and normal, accounting for approximately 15\%, 25\% and 60\% of the test set, respectively. Evaluations are conducted on different image degradation sequences, and the results are shown in Table \ref{tab:difference_scenes}. It can be seen that in scenarios with severe image degradation, e.g., dark and dazzle, MSSF-PP's improvements over MSSF-PP-R is relatively small (+10.5\% and +2.8\% on $\text{mAP}_\text{BEV}$). As the image quality improves (i.e., normal scenario), the gain also increases (+15.6\% on $\text{mAP}_\text{BEV}$). This conclusion is obvious because images cannot provide accurate object information in bad scenarios. However, it is worth noting that our method does not produce negative gains due to image degradation, which shows that our fusion method has a certain degree of robustness.

\begin{figure*}[!t]
    \centering
    \includegraphics[width=7in]{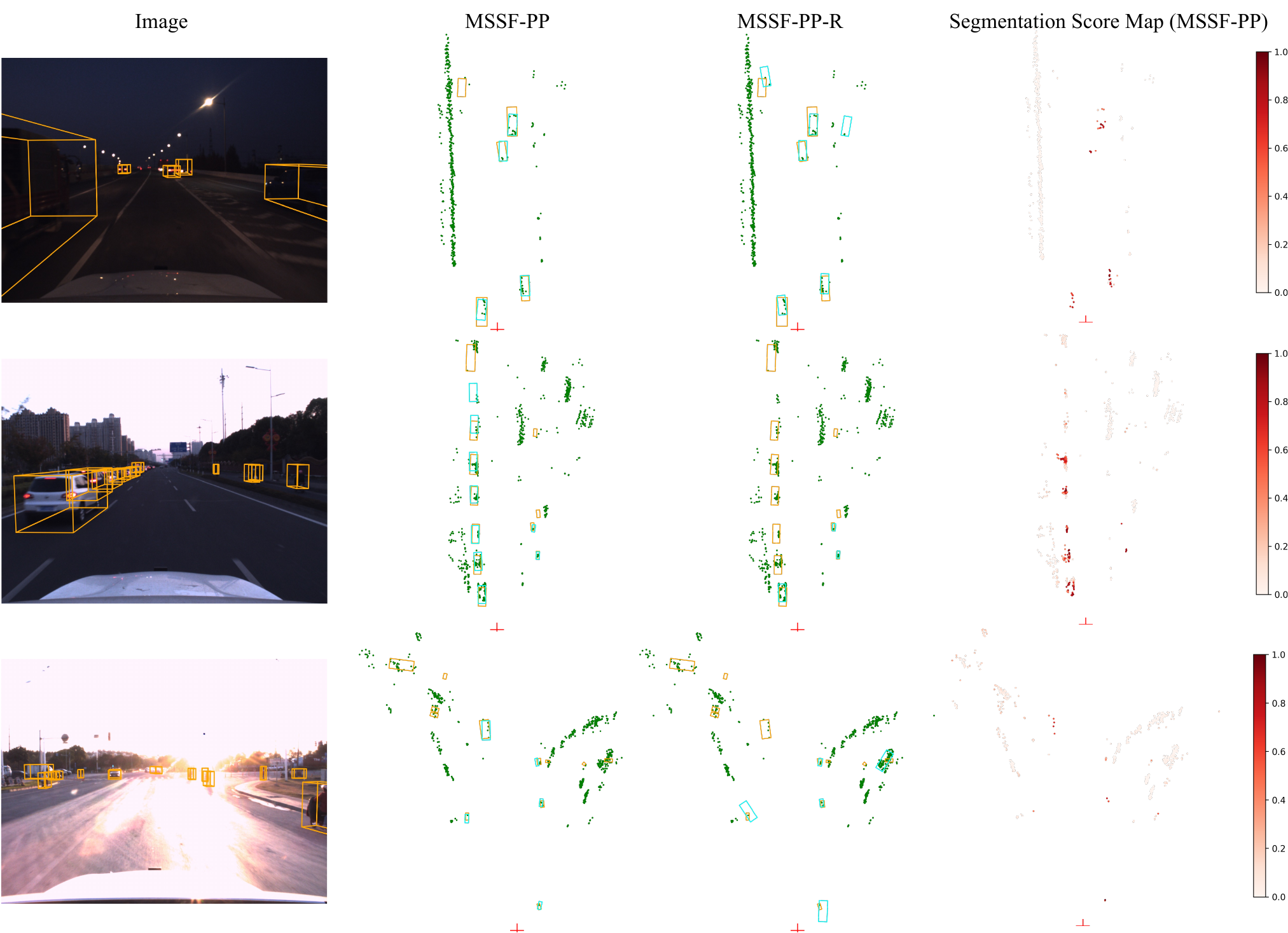}
    \caption{Visualization results on the TJ4DRadset dataset. The content and meaning are consistent with Fig. \ref{fig:vis_vod}.}
    \label{fig:vis_tj4d}
\end{figure*}

\paragraph{Visualization} Fig. \ref{fig:vis_tj4d} shows the visualization results of MSSF-PP and MSSF-PP-R on the TJ4DRadSet dataset. The three rows show the dark, normal and dazzle conditions. Even when the image quality is severely degraded, our method still shows robust detection results and is not affected by the failure of image features. Under normal lighting conditions, our method steadily improves the performance of the single-modal baseline, as shown in the second row. The fourth column demonstrates the ability of our method to distinguish 3D foreground points, even under dark conditions.

\begin{table}[!t]
    \caption{The Performance in Different Scenarios\label{tab:difference_scenes}}
    \centering
    \resizebox{\linewidth}{!}{
    \begin{tabular}{c|ccc|ccc}
    \hline  \multirow{2}{*}{Model}  & \multicolumn{3}{c|}{ $\text{AP}_{\text{3D}}$ (\%) } & \multicolumn{3}{c}{ $\text{AP}_{\text{BEV}}$ (\%) }\\
    \cline{2-7}
        & Dark & Dazzle & Normal & Dark & Dazzle & Normal\\
    \hline 
    MSSF-PP-R & 19.97 & 16.85 & 33.15 & 23.79 & 33.12 & 39.09 \\
    MSSF-PP  & \textbf{31.77} & \textbf{28.71} & \textbf{48.47} & \textbf{34.32} & \textbf{35.93} & \textbf{54.66}\\
    \hline
    \end{tabular}
    }
\end{table}

\subsubsection{The Plug-and-Play Capability of MSSF}

\begin{table}[!t]
    \caption{The Plug-and-Play Capability of MSSF\label{tab:plugandplay}}
    \centering
    \resizebox{\linewidth}{!}{
    \begin{tabular}{c|ccc|c|ccc|c}
    \hline  \multirow{2}{*}{Method}  & \multicolumn{4}{c|}{ $\text{AP}$ in the EAA (\%) } & \multicolumn{4}{c}{ $\text{AP}$ in the DC (\%) }\\
    \cline{2-9}
        & Car & Ped. & Cyc.  & mAP & Car & Ped. & Cyc.  & mAP\\
    \hline 
    CenterPoint\cite{yinCenterbased3DObject2021}  
    & 35.84 & 41.03 & 67.11 & 47.99 & 70.65 & 50.14 & 85.67 & 68.82 \\
    \textit{+MSSF} 
    & 52.13 & 49.28 & 76.97 & 59.46\scriptsize{\,(+11.47)} & 81.07 & 59.54 & 87.31 & 75.97\scriptsize{\,(+7.15)} \\
    \hline
    SECOND\cite{yanSECONDSparselyEmbedded2018a}
    & 40.95 & 39.97 & 64.92 & 48.61 & 72.05 & 50.25 & 81.99 & 68.10 \\
    \textit{+MSSF} 
    & 53.92 & 50.97 & 75.49 & 60.13\scriptsize{\,(+11.52)} & 90.44 & 60.91 & 94.16 & 81.84\scriptsize{\,(+13.74)} \\
    \hline
    VoxelNeXt-2D\cite{chenVoxelNeXtFullySparse2023a}   
    & 38.34 & 41.62 & 68.07 & 49.35 & 71.48 & 52.38 & 87.68 & 70.51 \\
    \textit{+MSSF} 
    & 52.48 & 50.69 & 77.87 & 60.35\scriptsize{\,(+11.00)} & 80.90 & 60.55 & 88.66 & 76.70\scriptsize{\,(+6.19)} \\
    \hline
    \end{tabular}
    }
\end{table}

To demonstrate the plug-and-play capability of our method and validate the effectiveness of the proposed fusion framework, we integrate MSSF with three detection networks: CenterPoint\cite{yinCenterbased3DObject2021}, SECOND\cite{yanSECONDSparselyEmbedded2018a}, and VoxelNeXt-2D\cite{chenVoxelNeXtFullySparse2023a}. CenterPoint and SECOND are representative voxel-based methods, while VoxelNeXt-2D is the pillar version of VoxelNeXt\cite{chenVoxelNeXtFullySparse2023a}. Table \ref{tab:plugandplay} summarizes the detection performance of these networks on the VoD validation set, both with and without MSSF. The results highlight consistent and significant performance improvements across all networks after incorporating MSSF. This demonstrates that our fusion block can effectively capture and fuse image features into different detection networks, leading to enhanced detection performance.

\subsection{Ablation Experiments}
To analyze and verify the effectiveness of the proposed fusion network, we conduct detailed ablation experiments on the VoD dataset. The following experiments are done based on MSSF-V. In addition, the VoD dataset adopts 11-point sampling when calculating AP. However, we find that the AP calculated by 11-point sampling fluctuates more than that of 40-point sampling, especially when the recall rate is near the sampling point. As a result, for better analysis, we present the results in $\text{AP}_{40}$, which is widely adopted in datasets like KITTI\cite{geigerAreWeReady2012}, for some experiments below.

\begin{figure}[!t]
    \centering
    \subfloat[][Blur Ratio]{\includegraphics[width=1.7in]{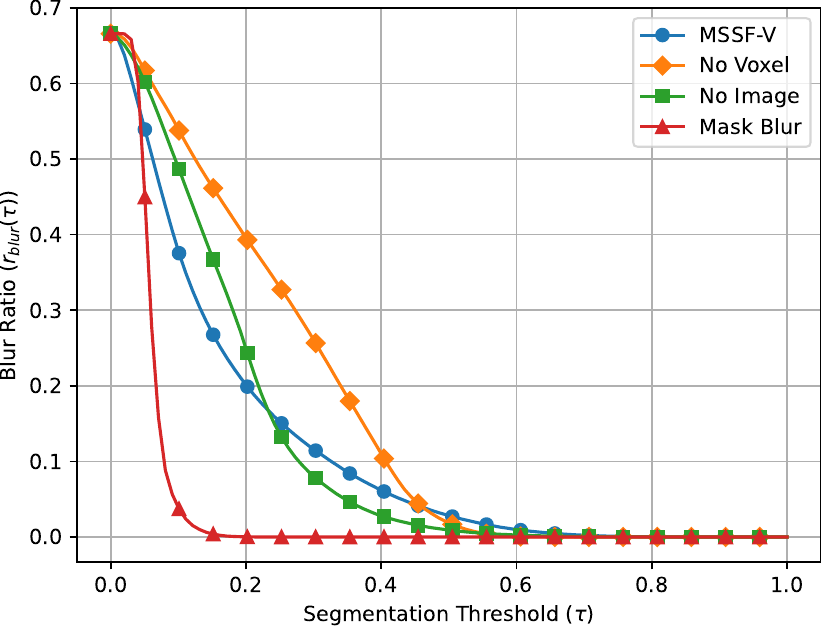}
    \label{fig:br}}
    \hfil
    \subfloat[][Foreground Recall]{\includegraphics[width=1.7in]{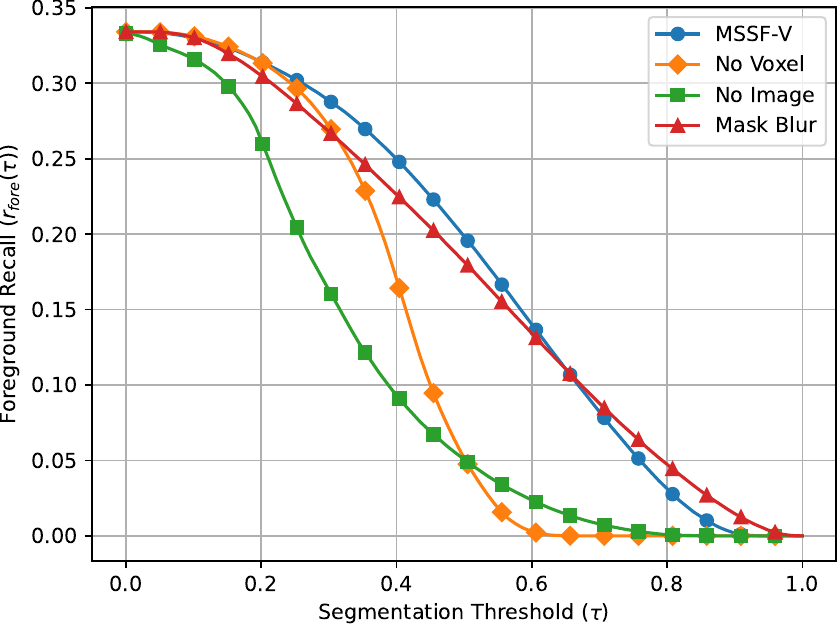}%
    \label{fig:fr}}
    \caption{The blur ratio (lower is better) and foreground recall (higher is better) curve with respect to the segmentation threshold $\tau$.}
    \label{fig:brfr}
\end{figure}

\begin{table}[!t]
    \caption{Effects of the Feature-Blurring Problem\label{tab:feature_blurring}}
    \centering
    \resizebox{\linewidth}{!}{
    \begin{tabular}{c|ccc|c|ccc|c}
    \hline  \multirow{2}{*}{Method}  & \multicolumn{4}{c|}{ $\text{AP}$ in the EAA (\%) } & \multicolumn{4}{c}{ $\text{AP}$ in the DC (\%) }\\
    \cline{2-9}
        & Car & Ped. & Cyc.  & mAP & Car & Ped. & Cyc.  & mAP\\
    \hline 
    No Voxel & 46.99 & 33.05 & 53.00 & 44.35 & 80.89 & 42.11 & 71.26 & 64.75\\
    No Image & 37.12 & 46.24 & 74.34 & 52.56 & 72.18 & 53.01 & 87.78 & 70.99\\
    MSSF-V & 50.10 & 50.92 & 77.57 & 59.53 & \textbf{81.07} & 61.59 & 88.26 & 76.98 \\
    Mask Blur & \textbf{50.75} & \textbf{56.37} & \textbf{79.95} & \textbf{62.36} & 80.42 & \textbf{68.36} & \textbf{90.91} & \textbf{79.90}\\
    \hline
    \end{tabular}
    }
\end{table}

\begin{table*}[!t]
\caption{Effects of Different Fusion Blocks and \# Fusion Stages\label{tab:fusion_block}}
\centering
\begin{tabular}{c|c|ccc|cc|ccc|cc|c}
\hline \multirow{2}{*}{Fusion Block} & \multirow{2}{*}{$n$} & \multicolumn{5}{c|}{ AP in the Entire Annotated Area (\%) } & \multicolumn{5}{c|}{ AP in the Driving Corridor (\%) } & \multirow{2}{*}{FPS}\\
\cline{3-12}
    &  & Car & Pedestrian & Cyclist & mAP  & $\text{mAP}_\text{str}$ & Car & Pedestrian & Cyclist & mAP & $\text{mAP}_\text{str}$ \\
\hline 
\multirow{5}{*}{MSDFF}  
& 0 & 38.28 & 42.93 & 69.96 & 50.39 & 27.53 & 71.76 & 52.92 & \textbf{88.93} & 71.21 & 48.59 & \textbf{24.6}\\
& 1 & 50.10 & 50.92 & 77.57 & 59.53 & 37.30 & 81.07 & 61.59 & 88.26 & 76.98 & 63.88 & 10.7\\
& 2 & 52.53 & 51.58 & 75.77 & 59.96 & 39.92 & 89.08 & \textbf{66.78} & 88.10 & \textbf{81.32} & 66.44 & 10.3\\
& 3 & \textbf{53.50} & 52.22 & 74.55 & 60.09 & 40.82 & \textbf{90.85} & 61.66 & 87.61 & 80.04 & 65.84 & 9.9\\
& 4 & 53.49 & \textbf{51.66} & 69.79 & 58.32 & 41.07 & 90.59 & 61.86 & 88.40 & 80.28 & 66.11 & 9.5\\

\hline
\multirow{5}{*}{SFF} 
& 0 & 38.28 & 42.93 & 69.96 & 50.39 & 27.53 & 71.76 & 52.92 & \textbf{88.93} & 71.21 & 48.59 & \textbf{24.6}\\
& 1 & 46.62 & 49.81 & 75.46 & 57.30 & 35.46 & 80.89 & 60.90 & 87.69 & 76.49 & 61.01 & 10.9\\
& 2 & 53.20 & 51.31 & \textbf{76.18} & 60.23 & 41.78 & 81.66 & 61.77 & 87.37 & 76.93 & 64.79 & 10.5 \\
& 3 & 53.23 & 51.58 & \textbf{76.18} & \textbf{60.33} & \textbf{41.79} & 90.16 & 62.14 & 87.32 & 79.88 & 64.52 & 10.2\\
& 4 & 53.27 & 51.39 & 69.70 & 58.12 & 40.85 & 90.08 & 61.44 & 87.66 & 79.73 & \textbf{66.60} & 9.8\\
\hline
\end{tabular}
\end{table*}

\paragraph{Effects of the Feature-Blurring Problem}
We investigate the mitigating effect of our model on the feature-blurring problem and explore the impact of feature-blurring on detection performance. For analysis, we categorize the centroids of non-empty voxels that are input to the semantic-guided head into 2D foreground points, 3D foreground points, and 3D blurred points, following the definition in Fig. \ref{fig:motivate}(a), where the 2D instance masks in the definition are replaced by the 2D ground truth boxes. We further define the blur ratio as $r_{blur}(\tau) = \frac{n_{blur}(\tau)}{n_{fore2d}}$ and the foreground recall as $r_{fore}(\tau) = \frac{n_{fore3d}(\tau)}{n_{fore2d}}$, where $n_{fore2d}$ is the number of 2D foreground points, $n_{fore3d}(\tau)$ and $n_{blur}(\tau)$ is the number of 3D foreground points and the number of 3D blurred points above the segmentation threshold $\tau$, respectively. In Fig. \ref{fig:brfr}(a) and (b), we depict the curves of $r_{blur}(\tau)$ and $r_{fore}(\tau)$ with respect to $\tau$, respectively. In the legend, ``No Image'' means no use of image features in MSSF-V (equivalent to a radar-only version with the semantic-guided head). ``No Voxel'' signifies not using point cloud voxel features in MSSF-V, which means the voxel feature $\mathbf{f}_{vox}$ is excluded in the query initialization (i.e., Eq. (\ref{equ:init_query})) and the fusion operator (i.e., Eq. (\ref{equ:fusion_operator})). ``Mask Blur'' denotes masking out the image features sampled from 3D blurred points using ground truth boxes. Corresponding detection results are presented in Table \ref{tab:feature_blurring}.



As can be seen from Fig. \ref{fig:brfr}(a), when there is only image information (i.e., No Voxel), numerous 3D blurred points contain image features of foreground objects without geometric features. This makes them difficult to distinguish, leading to a high blur ratio. In Table \ref{tab:feature_blurring}, the corresponding detection performance also drops significantly, especially for pedestrians and cyclists. When only point cloud information is available (i.e., No Image), the 3D blurred points are not assigned incorrect image semantic features, eliminating the feature-blurring problem and resulting in a low blur ratio. However, sparse radar point clouds result in a weak representation of object geometric information, causing a decrease in the model's discriminative ability, and subsequently exhibiting low foreground recall in Fig. \ref{fig:brfr}(b) and poor detection performance in Table \ref{tab:feature_blurring}. 
In contrast, our multi-stage fusion structure enables deep interaction between image and point cloud features, resulting in the fused features with high discriminability. As a result, we achieve both high foreground recall and low blur ratio, mitigating the feature-blurring problem and resulting better detection performance shown in Table \ref{tab:feature_blurring}.
When further masking out the image features sampled from the blurred points (i.e., Mask Blur), the blur ratio is minimized, leading to a further improvement in detection performance. This indicates that the feature-blurring problem has a substantive impact on detection performance under radar modality.

\paragraph{Effects of Different Fusion Blocks and \# Fusion Stages}
We conduct two groups of experiments to observe the performance difference between the two proposed fusion blocks and the impact of the number of fusion blocks. In the first group, we adopt the MSDFF-based fusion block and fix $n+m=6$. The number of fusion blocks $n$ increases from 0 to 4. When $n=0$, it degenerates into a single-modal baseline. In the second group, we choose the fusion block based on SFF, and the other experimental settings are the same as those of the first group. The experiment results are shown in Table \ref{tab:fusion_block}. For better comparison, we also list the mAP calculated at stricter IoU thresholds, i.e., $\text{mAP}_{\text{str}}$. In particular, the IoU thresholds corresponding to the categories of cars, pedestrians, and cyclists are 0.7, 0.5, and 0.5, respectively. 

For the first group, the detection performance compared to the baseline has been significantly improved, when only one fusion block is used. When the number of fusion blocks gradually increases, mAP does not increase significantly. On the contrary, there is a slight decrease in AP for the cyclist category. This is because the rich image features make it easy for the network to mistake distant cyclists as pedestrians. For cyclists who are far away or blocked, the bicycle is easily confused with the background, and the rider above is easier to detect. In addition, the deeper the block, the larger the corresponding voxel volume, causing the image features sampled by its centroid no longer represent the entire voxel accurately, especially when $n=4$. However, it should be noticed that as $n$ increases, $\text{mAP}_{\text{str}}$ improves. This result indicates that sufficient image features and deep fusion can help object positioning. This makes sense for radar, as radar point clouds lack accurate geometric information of objects due to sparsity and noisiness. Nevertheless, information like object orientation is easier to perceive in images. More accurate object localization can be achieved through deep interaction of voxel and image features.
Furthermore, as the fusion block has higher computational costs than the ordinary block, FPS also decreases slightly, as $n$ increases. The decrease in FPS is slight thanks to the sparsity of radar point clouds.

Interestingly, the improvement for the second group is not as obvious as that in the first group when $n=1$. This is because the MSDFF has a stronger image feature extraction ability than SFF. The MSDFF can adaptively focus on the desired image features depending on the query. As $n$ increases (except $n=4$), both mAP and $\text{mAP}_\text{str}$ increase. In particular, the improvement of $\text{mAP}_\text{str}$ is obvious, which is consistent with the first group. When $n=2,3$, it even surpasses the first group under the EAA. This is because multi-stage feature fusion drowns out the advantages of MSDFF. In addition, the increase in voxel volume leads to unrepresentative image features as mentioned above, which also weakens the effect of MSDFF.

In summary, both groups of results illustrate the effectiveness of the proposed multi-stage sampling, especially under tighter IoU thresholds.

\paragraph{Effects of the Semantic-Guided Head}

\begin{table}[!t]
\caption{Effects of the Semantic-Guided Head\label{tab:seg_head}}
\centering
\resizebox{\linewidth}{!}{
\begin{tabular}{c|cc|ccc|c|ccc|c}
\hline  \multirow{2}{*}{$n$} & \multicolumn{2}{c|}{Operation}  & \multicolumn{4}{c|}{ $\text{AP}_{40}$ in the EAA (\%) } & \multicolumn{4}{c}{ $\text{AP}_{40}$ in the DC (\%) }\\
\cline{2-11} 
& SL & SW & Car & Ped. & Cyc.  & mAP & Car & Ped. & Cyc.  & mAP\\
\hline 
\multirow{3}{*}{1} 
&            &            & 47.44 & 44.43 & 75.91 & 55.92 & 84.40 & 53.84 & \textbf{94.20} & 77.48\\
&\checkmark &            & 49.60 & 48.66 & 73.53 & 57.26 & 88.68 & 55.58 & 91.62 & 78.63\\
&\checkmark & \checkmark & 48.65 & 49.09 & \textbf{76.26} & 58.00 & 83.96 & 60.61 & 94.05 & 79.54\\
\hline
\multirow{3}{*}{3} 
&            &            & 52.66 & 50.67 & 72.59 & 58.64 & 88.87 & \textbf{62.39} & 93.32 & 81.52\\
&\checkmark &            & 53.70 & \textbf{52.52} & 73.05 & 59.76 & 91.88 & 62.29 & 93.91 & 82.69\\
&\checkmark & \checkmark & \textbf{54.38} & 51.44 & 74.01 & \textbf{59.94} & \textbf{94.60} & 62.05 & 93.29 & \textbf{83.31}\\
\hline
\end{tabular}
}
\end{table}
    
We analyze the effect of two components in the semantic-guided head, namely, the segmentation loss (SL) and the segmentation score weighting (SW). Table \ref{tab:seg_head} shows the experimental results in $\text{AP}_{40}$ when $n=1$ and $n=3$. It is observed that when the semantic-guided head is not used, that is, neither SL nor SW is used, $\text{mAP}_\text{EAA}$ decreases by 2.1\% and 1.3\% for $n=1$ and $3$, respectively. Despite this, the detection performance without the semantic-guided head is still significantly better than the single-modal baselines. The reason is that the classification loss in the detection head plays a similar role to SL, which can help the model identify 3D foreground points based on the fused BEV feature map with radar and image information. When SL is added but SW is not used, the detection performance is improved by 1.3\% and 1.1\%, indicating that additional explicit supervision can promote the network's discrimination of foreground and background points. However, only explicit supervision cannot effectively utilize the segmentation results, and the predicted foreground scores cannot be fed back to the network. After adding SL and SW at the same time, the detection performance is further improved. This shows that it is beneficial for the network to explicitly perceive the foreground and background in radar point clouds, when the feature-blurring problem is serious.


\begin{table}[!t]
\caption{Effects of Different Fusion Locations\label{tab:fusion_loc}}
\centering
\resizebox{\linewidth}{!}{
\begin{tabular}{c|ccc|c|ccc|c}
\hline  \multirow{2}{*}{Location}  & \multicolumn{4}{c|}{ $\text{AP}_{40}$ in the EAA (\%) } & \multicolumn{4}{c}{ $\text{AP}_{40}$ in the DC (\%) }\\
\cline{2-9}
    & Car & Ped. & Cyc.  & mAP & Car & Ped. & Cyc.  & mAP\\
\hline 
AR & 53.72 & 51.38 & 72.48 & 59.19 & 91.84 & \textbf{62.06} & \textbf{94.15} & 82.68\\
BR & \textbf{54.38} & \textbf{51.44} & \textbf{74.01} & \textbf{59.94} & \textbf{94.60} & 62.05 & 93.29 & \textbf{83.31}\\
\hline
\end{tabular}
}
\end{table}


\begin{figure}[!t]
    \centering
    \subfloat[][Before the residual block (BR)]{\includegraphics[width=1.7in]{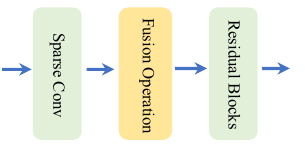}
    \label{fig:br_loc}}
    \hfil
    \subfloat[][After the residual block (AR)]{\includegraphics[width=1.7in]{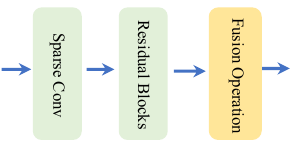}%
    \label{fig:ar_loc}}
    \caption{The structures of BR and AR.}
    \label{fig:brar}
\end{figure}

\paragraph{Effects of Different Fusion Locations}
As can be seen from Fig. \ref{fig:fusion_block}, the fusion process occurs between sparse convolution and residual blocks. The purpose of this design is to further process the fused features and interact with neighborhood features through residual blocks. In Table \ref{tab:fusion_loc}, we examine the impact of the fusion location, where ``BR'' represents fusion before the residual block, which is the default setting of our model, and ``AR'' represents fusion after the residual block. These two structures are illustrated in Fig. \ref{fig:brar}. It can be seen from Table \ref{tab:fusion_loc} that the two fusion locations have no obvious impact on the results. In particular, AR is slightly inferior to BR. This is because the AR only has one sparse convolution layer between two fusion operations, and does not perform sufficient feature mapping and exchange to better process the fused features.

\paragraph{Computational Requirements Analysis}

\begin{table}[!t]
\caption{Computational Requirements\label{tab:comput}}
\centering
\resizebox{\linewidth}{!}{
\begin{tabular}{c|c|c|c|c}
\hline
Model & FLOPs (G) & \# Params. (M) & FPS & $\text{mAP}_\text{EAA}$ (\%) \\
\hline
BEVFusion (ICRA 2023)\cite{liuBEVFusionMultiTaskMultiSensor2023}
& 385 & 39.8 & 7.1 & 49.25 \\
LXL (IEEE T-IV 2023)\cite{xiongLXLLiDARExcluded2023b}
& - & - & 6.1 & 56.31 \\
MSSF-PP
& 388 & 32.3 & 13.9 & 63.31 \\
\hline
\end{tabular}
}
\end{table}

Table \ref{tab:comput} presents the floating-point operations per second (FLOPs), parameter count, and inference speed of MSSF-PP, compared with BEVFusion\cite{liuBEVFusionMultiTaskMultiSensor2023} and LXL\cite{xiongLXLLiDARExcluded2023b}. Under the same hardware and input resolution, our method achieves higher inference speed and comparable FLOPs while delivering significantly better detection performance. Notably, the FLOPs of the BEV pooling operation in BEVFusion are not included, despite its non-negligible time cost, which reaches approximately 20ms at our resolution. In contrast, our method does not involve explicit view transformation but instead uses sampling, with MSDFF requiring only about 2ms under the same conditions.




\section{Conclusion}\label{sec:conclusion}
In this study, we proposed a simple but efficient 4D radar and camera fusion network, namely MSSF, for 3D object detection. Specifically, we designed two plug-and-play fusion blocks to effectively utilize and fuse image features, which alleviate the feature-blurring problem in image feature sampling. Through the cascade of multiple fusion blocks, the detection performance is further improved, especially in terms of the localization accuracy of bounding boxes, which is attributed to the deep interaction between voxel features and image features. 
Moreover, we devised a semantic-guided head to guide the network to explicitly perceive foreground points, thereby further mitigating the feature-blurring problem and enhancing performance. The effectiveness of our method has been verified on the VoD and TJ4DRadset datasets, significantly surpassing existing methods and achieving new state-of-the-art performance.
Notably, MSSF even outperforms the classic LiDAR-based models.

Note that our MSSF provides a concise and effective architecture based on the fusion of 4D radar and camera. It can inspire subsequent work and remind upcoming efforts to pay attention to the feature-blurring problem. Future work may involve leveraging temporal information and the depth information estimated from images for better matching radar points and image pixels, thus eliminating feature blurring and improving detection capabilities. 







\bibliographystyle{IEEEtran}
\bibliography{mssf_references.bib}

\end{document}